\documentclass[10pt,twocolumn]{article}

\usepackage{iccv}
\usepackage{times}
\usepackage{epsfig}
\usepackage{graphicx}
\usepackage{amsmath}
\usepackage{amssymb}
\usepackage{booktabs}
\usepackage{subfigure}
\usepackage[accsupp]{axessibility}  % Improves PDF readability for those with disabilities.
\usepackage[pagebackref=true,breaklinks=true,letterpaper=true,colorlinks,bookmarks=false]{hyperref}

\iccvfinalcopy % *** Uncomment this line for the final submission

\def\iccvPaperID{6433} % *** Enter the ICCV Paper ID here

\ificcvfinal\fi

\usepackage{multirow}
\usepackage[normalem]{ulem}
\usepackage[linesnumbered,ruled]{algorithm2e}
\usepackage{cleveref}
\usepackage{amsmath}
\usepackage{booktabs}
\usepackage{varwidth}
\usepackage{paralist}
\usepackage{enumerate}
\usepackage{amssymb}
\usepackage{pifont}
\usepackage{bm}
\usepackage{amsmath}
\usepackage{multirow}
\usepackage[dvipsnames]{xcolor}

\usepackage{caption}
\DeclareMathOperator*{\argmax}{arg\,max}

\crefname{equation}{Eq.}{Eq.}
\crefname{section}{Section}{Sections}
\crefname{subsection}{Section}{Sections}
\crefname{subsubsection}{Section}{Sections}
\crefname{figure}{Figure}{Figures}
\crefname{table}{Table}{Tables}
\crefname{subfigure}{Figure}{Figures}
\crefname{algocf}{Algorithm}{Algorithms}

\newcommand{\Trans}[1]{{#1}^{\top}}

\newcommand\blfootnote[1]{%
  \begingroup
  \renewcommand\thefootnote{}\footnote{#1}%
  \addtocounter{footnote}{-1}%
  \endgroup
}

\begin{document}

%%%%%%%%% TITLE
% \title{Finding Natural Language Concepts from Visual Datasets}
% \title{Searching Representative Visual Attributes in Natural Language}
% \title{Finding Representative Concepts for Explainable Image Classification}
% \title{Distill Labels into Discriminative Visual Features in Natural Language}
% \title{Learning Compact and Expressive Visual Representations in Natural Language}
% \title{Learning Concise and Expressive Visual Attributes in Natural Language}
\title{Learning Concise and Descriptive Attributes for Visual Recognition}
% \title{Learning Concise and Natural-Language Attributes for Visual Recognition}
% \title{Learning Compact and Expressive Visual Representations Explained by Natural Language}
% - What traditional representation learning does: Understanding -> Embedding in numbers -> Discriminative
% - What we did: Understanding -> Embedding in natural language -> Discriminative
% - That ICLR paper is also natura language embedding, but with many more dimensions
% \title{Pruning Visual Concepts for Explainable Image Classification}
% \title{Explainable Image Classification by Pruning Concepts for Language Bottleneck Models}

% \author{An Yan\\
% UC San Diego\\
% \and

\author{
% An Yan, Yu Wang, Yiwu Zhong, Chengyu Dong, Zexue He, Yujie Lu, William Wang, Jingbo Shang, Julian McAuley 
  An Yan$^{\ast\diamondsuit}$,
  Yu Wang$^{\ast\diamondsuit}$,
  Yiwu Zhong$^{\ast\clubsuit}$, 
  Chengyu Dong$^{\diamondsuit}$,
  Zexue He$^{\diamondsuit}$,
  \\
  Yujie Lu$^{\spadesuit}$,
  William Yang Wang$^{\spadesuit}$,
  Jingbo Shang$^{\diamondsuit}$, 
  Julian McAuley$^{\diamondsuit}$
  \\
  $^{\diamondsuit}$UC San Diego, 
  $^{\clubsuit}$University of Wisconsin-Madison,
  $^{\spadesuit}$UC Santa Barbara
  \\
  {\tt \small \{ayan,yuw164,cdong,zehe,jshang,jmcauley\}@ucsd.edu} \\
  {\tt \small \{yujielu,william\}@cs.ucsb.edu, yzhong52@wisc.edu}
  % {\small \{wanrongzhu,yujielu,wendaxu,migueleckstein,wangwilliamyang\}@ucsb.edu, ayan@ucsd.edu, xwang366@ucsc.edu}
}

% \and

\maketitle
\blfootnote{$^\ast$ equal contributions.}
\ificcvfinal\thispagestyle{empty}\fi

% Questions:
% 1. learn a set of representative attributes for any task (dataset). How to write the story?
% 2. How do we explain the problem when K is large, random words == gpt-3 attributes? Why does this finding indicate cases with a large K is meaningless?
% 2.1 Why random words is : random words vs 512 gpt-3 attributes vs 512 similar words 
% 3. Figure-2. Do we need to add 128-256?
% 4. Let model provide explainations to the decision process. Previously it human who teach the model, now we want LLM teach human
% 5. Summarize knowledge for human. 

%%%%%%%%% ABSTRACT
\begin{abstract}
   Recent advances in foundation models present new opportunities for interpretable visual recognition -- one can first query Large Language Models (LLMs) to obtain a set of attributes that describe each class, then apply vision-language models to classify images via these attributes.
   Pioneering work shows that querying thousands of attributes can achieve performance competitive with image features. 
   However, our further investigation on 8 datasets reveals that LLM-generated attributes in a large quantity perform almost the same as random words. 
   This surprising finding suggests that significant noise may be present in these attributes. 
   We hypothesize that there exist subsets of attributes that can maintain the classification performance with much smaller sizes, and propose a novel learning-to-search method to discover those concise sets of attributes. 
   As a result, on the CUB dataset, our method achieves performance close to that of massive LLM-generated attributes (e.g., 10k attributes for CUB), yet using only 32 attributes in total to distinguish 200 bird species. 
   Furthermore, our new paradigm demonstrates several additional benefits: higher interpretability and interactivity for humans, and the ability to summarize knowledge for a recognition task. 

\end{abstract}
\vspace{-10pt}

\section{Introduction}
Explaining black-box neural models is a critical research problem. 
For visual recognition, one line of research tries to classify objects with descriptions or attributes~\cite{farhadi2009describing,cimpoi2014describing,romera2015embarrassingly,kim2018interpretability,koh2020concept}, which provide additional information beyond visual cues such as activation maps~\cite{selvaraju2016grad, selvaraju2017grad}. However, they require in-depth human analysis and intensive annotation to obtain key attributes for a particular recognition task. Such a paradigm is costly and thus impractical to scale up when the number of classes and domains grows. 

The recent advance of language foundation models creates new opportunities for building interpretable visual recognition models, as demonstrated by the powerful capabilities of models such as GPT-3 and ChatGPT in encoding world knowledge~\cite{brown2020language,openai2022chatgpt,kocon2023chatgpt}.
One can query useful visual attributes from LLMs and classify images via these attributes by converting visual features from vision-language models (VLMs) (\eg, CLIP~\cite{radford2021learning}) into attribute scores~\cite{compDL}.
% A set of visual attributes that are related to visual classes can be effortlessly queried from LLMs. Then we can noisily label the correlation between an image and an attribute using VLMs (e.g., CLIP) by computing their similarity. Given a set of attributes, we can construct a semantic vector where each dimension corresponds to a visual attribute and the value represents the similarity between the image and the attribute. 
One recent work~\cite{yang2022language} shows that a large set of attributes from LLMs (\eg, ~50 attributes per class) can achieve comparable performance to image features in a linear probing setting. However, two key observations motivate us to re-think this formulation: (1) 
% enormous 
A large number of attributes dramatically hurts the interpretability of a model. It is unrealistic to manually check thousands of attributes to fully understand model decisions. (2) We surprisingly find that when the number of attributes is large enough (\eg, the dimension of image features), random words drawn from the entire vocabulary can perform equally well as LLM-generated attributes. Moreover, reducing the number of random words by 25\% can still attain competitive performance. 
This indicates that redundant and noisy information exists in the massive LLM-generated attributes.

\begin{figure}[t]
  \centering 
  \includegraphics[width=1\linewidth]{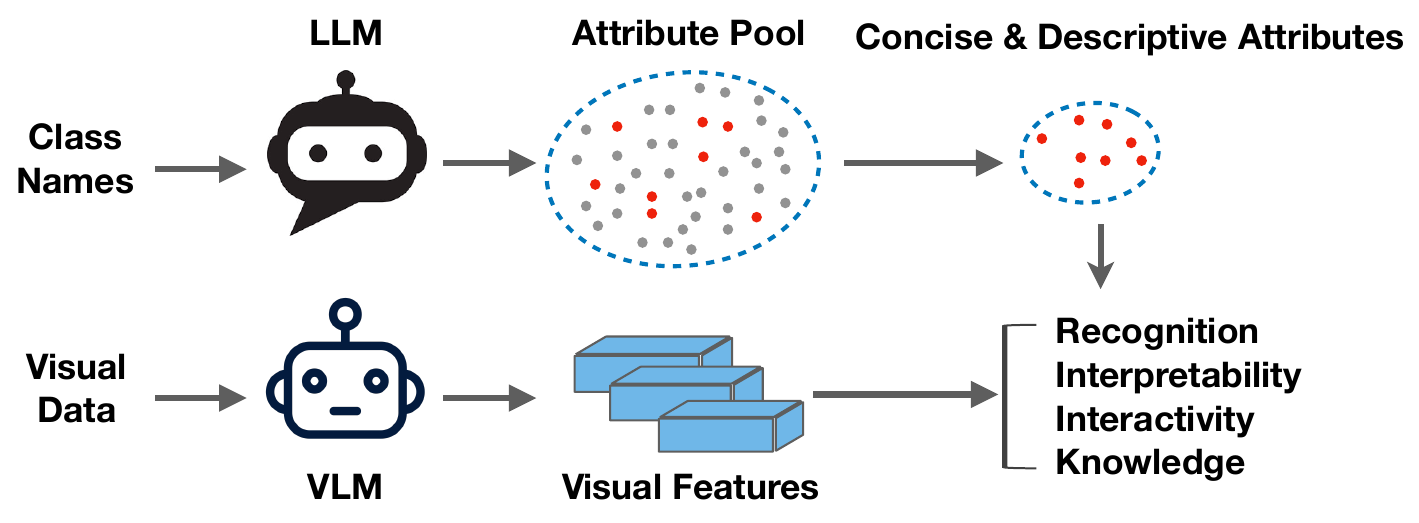}  
  \caption{Our proposed paradigm for visual recognition via learning a concise set of descriptive attributes. 
  %(LLM: large language model; VLM: vision-language model)
  }
  \label{fig:intro}
  \vspace{-10pt}
\end{figure}

% our work
With our findings, we ask the research question: \textit{Can we learn a concise set of representative visual attributes in the form of natural language to explain how visual recognition works?} \textbf{For example, can we find a few representative attributes to distinguish 200 bird species?}
% However, it is non-trivial to identify which visual attributes are discriminative for visual recognition. 
This is a non-trivial problem.
% Though LLMs are strong knowledge bases, they are not designed for visual recognition. There is still a gap to understand the visual world by learning only from text.
Even for humans, it is not easy to summarize what are the representative visual attributes given many visual classes.
To tackle this challenge, we propose a novel learning-to-search method, which uses image-level labels to guide the searching of discriminative attributes. Specifically, we train a learnable dictionary to approximate the embedding space of VLMs, and then find  descriptive attributes in the latent text space via  nearest neighbor search. 
% Our method follows a minimalist design and can be easily adapted to other visual recognition tasks beyond image classification in a plug-and-play fashion (\eg, object detection).
% with task-specific supervision to 

% in general --> in summary
In summary, we propose a new paradigm for visual recognition~(\cref{fig:intro}), which seeks to learn a concise set of visual attributes in the form of natural language. Once learned, there are several benefits to our new paradigm:
\textbf{(1)}~Our discovered attributes are highly descriptive. On 8 visual recognition datasets, our model classifies images via these attributes and achieves comparable classification performance as image features, even if the number of attributes is much smaller than the dimension of image features.
\textbf{(2)}~The condensed sets of attributes enable strong interpretability for the model decision process through a few human-friendly text descriptions.  
\textbf{(3)}~Additionally, our framework presents a natural language interface for humans to interact with. One can correct a wrong prediction during model inference, by perturbing the values of attribute scores where it made mistakes.
\textbf{(4)}~Lastly, these expressive attributes can be viewed as a concise form of knowledge to summarize useful features for a visual recognition task, without costly human effort.

% \textbf{(4)}~Lastly, as we automatically summarize distinctive features for visual recognition, these descriptive features can be viewed as a concise form of knowledge to help human understand a task.

% (4) these concise sets of expressive attributes can be viewed as a form of knowledge to summarize hundreds of classes from a visual recognition task, which is costly for humans to conclude manually.
% \textbf{(2)}~These concise sets provide strong interpretability where the model decision can be explained by a few human-friendly text descriptions.  
% Lastly, our paradigm can condense knowledge from a given arbitrary visual recognition dataset into a few visual attributes in natural language. Humans can learn what are the distinctive attributes to classify hundreds of classes. Such knowledge is costly to summarize manually. 

Overall, our contributions are three-fold:
\vspace{-8pt}
\begin{itemize}
    \item Leveraging  recent advances in foundation models, we propose a new paradigm for visual recognition by learning a concise set of attribute descriptions. \vspace{-8pt}
    \item To find these attributes, we propose a novel learning-to-search method which prunes the large attribute pool from large language models to a descriptive subset. \vspace{-8pt}
    \item We conduct extensive experiments across 8 visual recognition datasets to validate our recognition effectiveness and efficiency with additional benefits.
\end{itemize}

\section{Methodology}
\label{sec:methodology}
\begin{figure*}
\centering
\includegraphics[width=\textwidth]{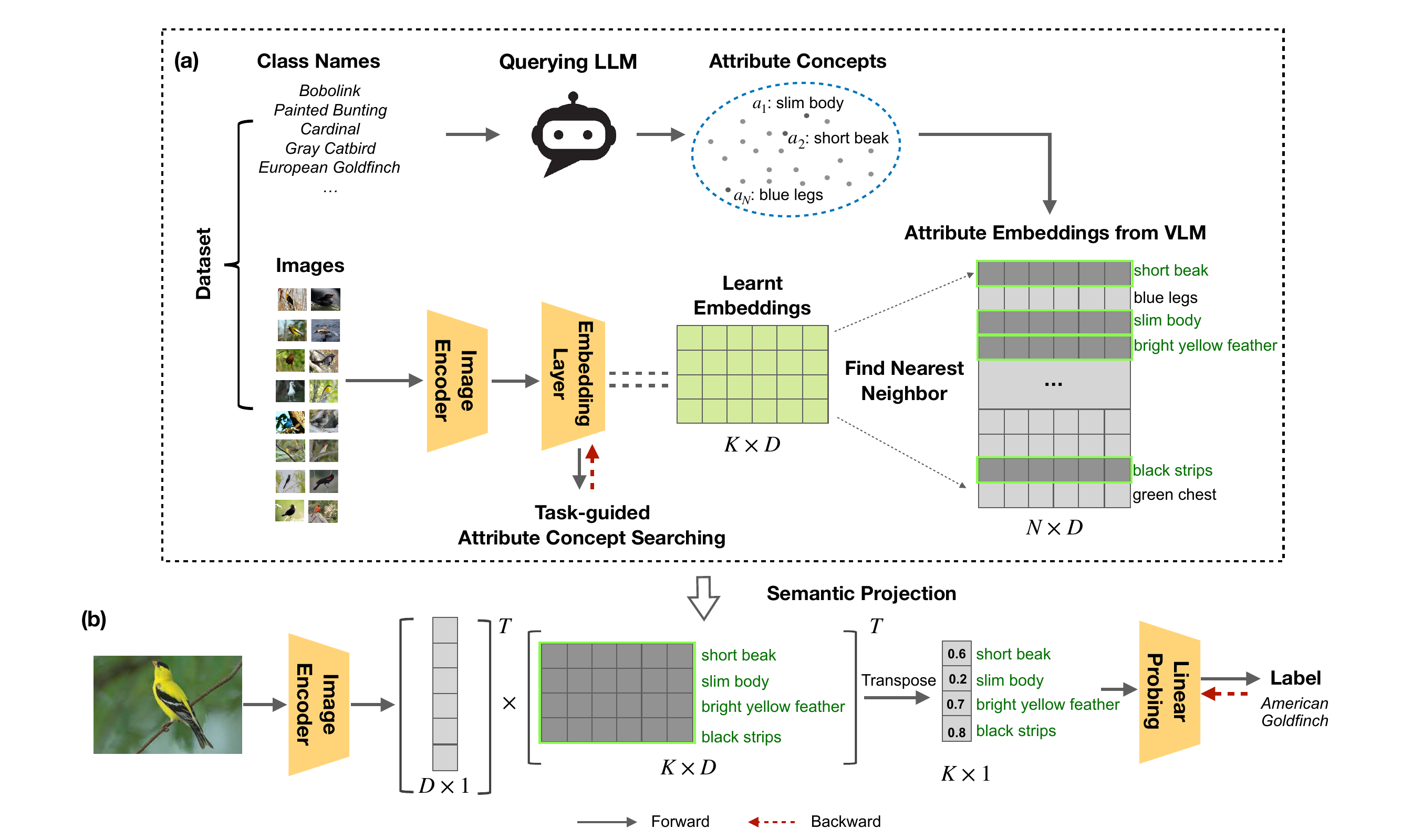}
\caption{The framework of our model. (a) Querying attributes from LLMs and finding a concise set of representative attributes; (b) An example using the attributes for interpretable visual recognition.}
% \vspace{-8pt}
\label{fig:model}
\end{figure*}

In this section, we introduce our key components for a new paradigm of visual recognition. It mainly consists of three modules: \textbf{First}, in~\cref{sec:method_llm}, given an image domain, we query large language models to obtain a large set of visual attributes for the categories of a task. \textbf{Second}, we use a semantic transformation (\cref{sec:method_projection}) to project the image features into attribute features via a vision-language model, where each dimension in the new space corresponds to an attribute concept, and a higher value represents higher correlation between the image and the attribute. 
\textbf{Finally}, given the large space of attributes, we propose a novel learning-to-search method (\cref{sec:method_searching}) to efficiently prune the attributes into a much smaller subset to obtain a concise model for classification.

\subsection{Generating Attribute Concepts via LLMs}
\label{sec:method_llm}
The first step of our framework is to obtain a set of appropriate attribute concepts. Given a dataset with different categories, (e.g.,~CUB with 200 bird classes), what are the distinctive visual attributes to recognize them? 
Manually labeling and designing these attribute concepts can be costly, and can not scale to large numbers of classes.  Large Language Models (LLMs), such as GPT-3~\cite{brown2020language} and ChatGPT, provide an alternative solution. We can view these language models as implicit knowledge bases with exceptional world knowledge on a variety of tasks and topics, which humans can easily interact with through natural language to query knowledge. To this end, prompt engineering, or the ability to ask good questions to language models, is still important. To effectively query knowledge from LLMs with regard to classifying images, we design two types of prompts.

\noindent \textbf{Instance Prompting for Class-level Features.} 
For each class $c$ in a given task, our first design choice is to query class-level information from LLMs. We prompt a language model with the instance prompt: 

\textit{Q: What are the useful visual features to distinguish $Y_c$ in a photo?}
% \yj{this is a Template; maybe there is a need to add robustness check in supplemental materials}
         
\noindent where $Y_c$ corresponds to the name of class $c$ in the form of natural language.

\noindent \textbf{Batch Prompting for Group-level Features.} 
For certain datasets (e.g., CIFAR-100 and ImageNet), there is inherently a hierarchy that some categories belong to the same group. For example, in CIFAR-100, there is a superclass for every five categories. Hence, we propose batch prompting, where we ask the language model to reason about the distinctive visual features among a batch of categories:

\textit{Q: Here are $N_g$ kinds of $Y_{g}$: \{$Y_{c_1}$,  $Y_{c_2}$, \ldots,  $Y_{c_M}$\}. What are the useful visual features to distinguish them in a photo?}

\noindent where $N_g$ is the number of classes in a group $g$, $Y_{g}$ is the name of the group,  $Y_{c_i}$ corresponds to the name of each class $c_i$ in the form of natural language.

We present more details regarding our prompt design, robustness check of different prompts, and examples of the generated attributes in Appendix~\ref{app:prompt_design}.
%Comparing with human written attribute concepts, attribute concepts generated by language models. 

\subsection{Semantic Projection}
\label{sec:method_projection}
After obtaining a pool consisting of $N$ attribute concepts $\mathcal{C} = \{a_1, a_2, \ldots, a_N\}$, the second challenge is how we can best leverage these attributes to build interpretable image classifiers. Recent advances of vision-language models such as CLIP bridge the gap between images and text, by pre-training models with large scale image-text pairs.
Intuitively, converting from images to text is a discretization process that will unavoidably lose rich semantic information stored in an image. 

To better preserve information, we use a semantic projection that transforms a visual feature into an attribute concept space. Given an image $I$, we convert the D-dimensional image feature $\textbf{V} \in \mathbb{R}^D$ into an N-dimensional attribute concept vector $\textbf{A} \in \mathbb{R}^N$:
% \begin{align}
%     \textbf{V} &= \Theta_V(I), \textbf{T}_i = \Theta_T(a_i) \label {eq:vlm} \\
%     s_i &= \cos(\textbf{V},  \textbf{T}_i), i=1, ..., N \label{eq:score}\\ \label{eqn:semantic_projection}
%     \textbf{A} &= (s_1, ..., s_N)^T
% \end{align}
\begin{align}
    \textbf{V} &= \Theta_V(I), \textbf{T}_i = \Theta_T(a_i) \nonumber \\
    s_i &= \cos(\textbf{V},  \textbf{T}_i), i=1, ..., N \label{eqn:semantic_projection}\\ 
    \textbf{A} &= (s_1, \ldots, s_N)^T \nonumber
\end{align}
where $\cos(\cdot,\cdot)$ is the cosine similarity between two vectors, $s_i$ is the cosine similarity between two vectors. $\Theta_V$ and $\Theta_T$ are the visual and text encoder of a VLM. $\textbf{T}_i$ is the embedding of the $i$-th attribute in the attribute concept pool, $i\in\{1,\ldots, N\}$. $\textbf{A}$ is the semantic vector of image $I$. 

\subsection{The Hypothesis of Attribute Concept Space}
\label{sec:method_hypothesis}
 Conceptually, our semantic projection resembles principal component analysis, where we aim to find a set of bases in the form of natural language, and by projecting the images into these bases we obtain a new attribute concept space where each dimension in the space corresponds to a visual attribute concept. 
However, the large bag of attribute concepts we obtained from large language models is not the optimal language basis. As of today, LLMs are models that noisily condense world knowledge from the web, and are not optimized for visual recognition or visual reasoning tasks. We hypothesize that there exist subsets of attributes that can still achieve high classification performance with a much smaller size.
Intuitively, most attributes in the large attribute concept pool are irrelevant to classify a certain class. For example, attributes that describe dogs are less likely to be suitable attributes to recognize birds or cars. 
Practically, formatting a compact attribute set is also helpful for humans to interact with the model and understand its behavior better. A small number of attributes is much easier for diagnostic purposes and making decisions with these neural models, which is the ultimate goal of building interpretable models.

\subsection{Task-Guided Attribute Concept Searching}
\label{sec:method_searching}
Finding an expressive set of language bases is non-trivial. The massive attributes from LLMs are noisy, and finding a few representative attributes for hundreds of classes in a task can be challenging and costly, even for human experts with domain knowledge.
An exhaustive search is also impractical given the large text space.
% Given $N$ attribute concepts, finding a subset of $K$ attributes ($K\ll N$) to achieve optimal classification performance is essentially a searching problem: the brute force solution is to exhaustively train different models from all possible combinations and find the one with the best performance, which is impractical due to high computational cost given the large search space.

Inspired by dictionary learning and vector quantization techniques~\cite{van2017neural}, we present a learning-to-search method that learns a dictionary to approximate an expressive subset of attributes given fixed $K$. 
Specifically, we first define an embedding matrix $\textbf{E} \in \mathbb{R}^{K\times D}$, 
% that should also be in text embedding space, 
where $K$ is a $K$-way categorical that equals the number of attributes, and $D$ is the dimensionality of embedding vectors $\textbf{V}$ and $\textbf{T}_i$ (\ie, the latent dimension of VLMs), where $\textbf{V}$ and $\textbf{T}_i$ is the image embedding and the $i$-th attribute embedding shown in Eq.(\ref{eqn:semantic_projection}). 
Since our goal is to find $K$ attributes to be expressive, we propose a task-guided attribute concept searching method to optimize for a particular task. For visual recognition tasks, we use a classification head to project the dictionary into $K_C$ classes and guide the learning process with the categorical cross-entropy loss:
\begin{equation}\label{eq:loss_cc}
    \mathcal{L}_{\mathit{ce}} = -\frac{1}{M} \sum_{i=1}^M \sum_{c=1}^{K_C} y_{i,c} \log(p_{i,c})
\end{equation}
where $M$ is the number of images in a mini-batch, $y_{i,c}$ is the binary indicator of the $i$-th image in the mini-batch belonging to class $c$, and $p_{i,c}$ is the predicted probability of the $i$-th image belonging to class $c$.

But simply training with the guidance of the cross-entropy loss is suboptimal, as the embeddings $\textbf{E}$ are not in the same space of $\textbf{T}$. Thus, we use the Mahalanobis distance as a constraint to encourage the embeddings to be optimized towards the latent space of vision-language models. Given a sampled probability distribution $\textbf{T}$, the Mahalanobis distance of $\textbf{E}_j$ from $\textbf{T}$ is defined as 
\begin{equation}
    \mathcal{D}_{\mathit{mah}}^j = \sqrt{(\textbf{E}_j - \boldsymbol{\mu}) \textbf{S}^{-1} (\textbf{E}_j - \boldsymbol{\mu})}
\end{equation}
where $\boldsymbol{\mu} = (\mu_1, ..., \mu_D)$ is the mean vector and $\textbf{S}$ is the positive-definite covariance matrix of $\textbf{T}$. Then the regularization term is defined as: 
\begin{equation}\label{eq:mahalanobis}
    \mathcal{L}_{mah}^j = \frac{1}{K}\sum_{j=1}^k \mathcal{D}_{\mathit{mah}}^j 
\end{equation}
Overall, our model is optimized with a mixture of two losses:
\begin{equation}\label{eq:loss_func}
    \mathcal{L}_{loss} = \mathcal{L}_{\mathit{ce}} +  \lambda \sum_{j=1}^K \mathcal{L}_{\mathit{mah}}^j .
\end{equation}
% Here $\lambda$ is a rescaling hyperparameter that weighs the two losses.

After training, we have the embedding matrix $\textbf{E}$ which will be used for searching the attributes from the attribute concept pool $\mathcal{C}$. Note that for $\textbf{E} \in \mathbb{R}^{K*D}$, each row of $\textbf{E}$ is a $D$-dimensional vector. We denote the $j$-th row of $\textbf{E}$ as $\textbf{E}_j$. We use greedy search as follows:
\begin{align}
    {\textbf{T}}^*_j &= \argmax_{i \in \{1,\cdots, N\}} \cos(\textbf{T}_i, \textbf{E}_j), \nonumber \\
    &\textrm{s.t. } {\textbf{T}^*_j\neq \textbf{T}^*_k, \forall 1\leq k<j}, \label{eq:text_embedding_searching} \\
    &\textrm{where } j \textrm{ is from $1$ to $K$}, \nonumber
\end{align}
As $j$ iterates from $1$ to $K$, we can find $K$ attribute embeddings $\textbf{T}^*_j, j\in\{1,\cdots,K\}$,
% which is stacked to matrix $\textbf{T}^* \in \mathbb{R}^{K*D}$. The attribute concepts 
which corresponds to $K$ expressive attribute concepts and are the condensed features containing the necessary knowledge for the task. With the selected attributes, we can calculate the semantic vector of each image as in~\cref{eqn:semantic_projection}, where each dimension of the vector is a similarity score between the image and an attribute. We evaluate the performance of these semantic vectors with linear probes, and the obtained linear model is used for inference and analysis.

% we find
% $K$ attribute concepts by conducting a nearest neighbour search using the learned embedding space $\textbf{E}$. 

% Classification guided attribute concept Searching

\section{Experiments}
% To illustrate the effectiveness of our method, we conduct extensive experiments to discuss and analyze our method in various aspects. We will start with the experimental setup, then present our main results, followed by ablation studies for each component, and finally the case studies to explore the interpretability and interactivitiy of our model in detail. 
\begin{table*}[ht]
    \centering
    \resizebox{0.95\linewidth}{!}{%
    \begin{tabular}{c|ccc|ccc|ccc|ccc}
    \toprule
        Datasets & \multicolumn{3}{|c|}{CUB} & \multicolumn{3}{c}{CIFAR-10} & \multicolumn{3}{|c}{CIFAR-100} & \multicolumn{3}{|c}{Flower}\\
        \midrule
        $K$ & 32 & 200 & 400 & 8 & 10 & 20 & 64 & 100 & 200 & 32 & 102 & 204   \\
        \midrule
        LaBo & -- & 60.93 & 62.61 & -- & 78.11 & 84.84 & -- & 75.10 &  76.94 & -- & 80.98 & 86.76 \\
        Ours & 60.27 & \textbf{63.88} & \textbf{64.05} & 77.47 & \textbf{80.09} & \textbf{87.99} & 73.31 &\textbf{75.12} & \textbf{77.29} & 80.88 & \textbf{87.26} &  \textbf{89.02} \\
        \midrule
        Datasets  & \multicolumn{3}{|c}{Food} & \multicolumn{3}{|c|}{Oxford\_Pets} & \multicolumn{3}{|c|}{Stanford\_cars} & \multicolumn{3}{|c}{Imagenet\_Animals} \\
        \midrule
        $K$ & 64 & 101 & 202 & 16 & 37 & 74  & 64 & 196 & 392  & 128 & 397 & 794 \\
         \midrule
         LaBo  & -- & 79.95 & 81.33 & -- & 76.91 & 84.33 &  -- & 72.33 & 74.39 & -- & 74.88 & 75.49 \\
        Ours  & 78.41 & \textbf{80.22} & \textbf{81.85} & 76.29 & \textbf{83.15} & \textbf{85.91} & 72.07 & \textbf{74.57} & \textbf{75.56} & 74.48 & \textbf{75.69} & \textbf{75.83}\\
        \bottomrule 
    \end{tabular}}
    \caption{Comparison with state-of-the-art. LaBo is designed to use at least as many attributes as classes. We use ``--" to denote non-applicability.}
\label{tab:comparison_with_labo}
\end{table*}

\subsection{Experimental Setup}
\paragraph{Datasets} We conduct our experiments on 8 different image classification datasets, including: CUB~\cite{cub}, CIFAR-10 and CIFAR-100~\cite{cifar10}, Food-101~\cite{food}, Flower~\cite{flower}, Oxford-pets~\cite{oxford_pets}, Stanford-cars~\cite{stanford_cars}, Imagenet~\cite{imagenet}. For Imagenet, it is not trivial to analyze all 1000 diverse classes. So we narrow the scope to 397 animal classes, with 509,230/19,850 samples for train/test. 
%samples out of 1,281,167 and 50,000 images, respectively. 
We denote this subset as Imagenet-Animals. For other datasets, most of them include images within a specific domain (CUB, Flower, Food, Oxford-pets, Stanford-cars), while CIFAR-10 and CIFAR-100 contain broader classes that lie across domains. 

% \vspace{-4pt}
\paragraph{Implementation Details}
Our method involves two stages of training. The first stage consists of task-guided learning of a dictionary $\textbf{E}$ to approximate CLIP text embeddings and using this dictionary to find $K$ attributes for visual recognition. For the Mahalanobis distance, the parameter $\lambda$ is tuned with a grid search in \{1, 0.1, 0.01, 0.001, 0\}. The second stage is one-layer linear probing to classify semantic vectors. 
The batchsize is set to 4,096 for all datasets except 32,768 on Imagenet-Animals for faster converging. We set the number of epochs to 5,000 epochs with early stopping. 
% \wy{chatgpt: The batch size has been configured to 4096 for all datasets, except for Imagenet-Animals, where it has been set to 32768 to expedite the convergence process. Additionally, we have designated a total of 5000 epochs with an early stopping mechanism.}
The learning rate is set to 0.01 in all experiments with an Adam optimizer~\cite{kingma2014adam}. Unless specified, we use GPT-3 and CLIP ViT-B/32 for all performance comparison.

% \vspace{-4pt}
\paragraph{Baselines}
We compare with state-of-the-art works that leverage attributes either from human annotations or from LLMs. For a fair comparison, we use linear probes to evaluate all methods: (1)~\textbf{CompDL}~\cite{compDL} builds semantic vectors using CLIP scores between human-designed attributes and images. (2)~\textbf{LaBO}~\cite{yang2022language} is a recent work that builds semantic vectors with a large set of attributes from LLMs. (3)~\textbf{Human}~\cite{cub,koh2020concept}. Attribute labels for each image are annotated by humans. We compare with two versions: binary labels for each attribute, and calibrated labels with confidence scores given by annotators.

To validate the effectiveness of learning-to-search, we explore other baselines: (1)~\textbf{K-means}. Perform K-means clustering on CLIP attribute embeddings, then find $K$ attributes with nearest distance to each clustering center. Intuitively this can be a strong baseline, as $K$ attributes close to each center can be distinctive.  (2)~\textbf{Uniform Sampling} from the large attribute pool.  (3)~\textbf{SVD}. After obtaining the attribute embeddings $\textbf{T}$, we run SVD decomposition of $\textbf{T}$ to get the top $K$ vectors and find attributes with the largest similarity with the $K$ important vectors. (4)~\textbf{Similarity}. We calculate the average score of each attribute across all images and then find the $K$ attributes with the largest average scores. (5) \textbf{Img Features}. Black-box linear probing on latent image features with two linear layers and an intermediate dimension $K$ as a reference.
% \subsection{Ablation study with Random Words}

% \begin{figure}
% \centering
% \subfigure[CUB]{\label{fig:cub}\includegraphics[width=0.480\linewidth]{figures/cub_scale.pdf}}
% \hfill
% \subfigure[Flower]{\label{fig:cifar10}\includegraphics[width=0.480\linewidth]{figures/flower_scale.pdf}}
% \caption{Performances with Random Words}
% \label{fig:overall_performances_with_random_words}
% \end{figure}

% \vspace{-8pt}
\subsection{Main Results}
\paragraph{Comparison with previous work}
We first compare our method with LaBo~\cite{yang2022language}. It is designed to use $M_c$ concepts per class with default number of 50, which corresponds to 10,000 attributes for CUB.
%and is way more than ours. 
For fair-comparison, we set $M_c$ as 1 and 2 in the experiments.
As shown in~\cref{tab:comparison_with_labo}, our method outperforms LaBo with the same number of attributes on both the full and few-shot setting. Furthermore, our method can achieve similar accuracy with only a smaller number of attributes (e.g., 32 attributes for CUB). These results suggest that our learned attributes are discriminative enough to classify the images, despite given much fewer attributes.

\begin{table}[]
    \centering
    \resizebox{1.0\linewidth}{!}
    {%
    \begin{tabular}{l|cccc}
    \toprule
        % & \multicolumn{3}{c}{K}\\
        % \midrule
        K (\# of attributes) & 8 & 16 & 32 & 312\\
        \midrule
        Human Binary~\cite{cub} & 4.02 & 7.31  & 10.11 & 47.38\\
        Human Calibration~\cite{koh2020concept}  & 3.75 & 7.15 & 9.78 & 43.37\\
        CompDL~\cite{compDL} & 12.64  &  26.41 &  28.69 & 52.60 \\
        Ours & \textbf{31.67}  & \textbf{48.55} & \textbf{60.27} & \textbf{65.17} \\
        % CLIP ViT-B/32 & 31.67  & 48.55 & 60.27 & 65.17 \\
        % CLIP ViT-L/14@336px  & 40.95 & 66.58  & 76.04 & 81.20\\
        \bottomrule 
    \end{tabular}}
    \caption{Comparison with human annotations on CUB.}
    \vspace{-10pt}
\label{tab:cub_human}
\end{table}
We then further compare with human annotations from CUB.  For $K<312$, we select attributes based on their accumulated confidence score for all samples. As shown in~\cref{tab:cub_human}, human annotated attributes are more noisy than CLIP similarities. With the same attributes, CLIP scores from CompDL build more expressive features. 
%On top of that, 
Furthermore,
our LLM-suggested attributes significantly outperform human designs, 
%it is able to use 
e.g.~by using
16 attributes we achieve similar performance as 312 attributes defined by humans.

\vspace{-4pt}
\paragraph{Large-scale attributes behave like random words}
\begin{figure}
    \centering
    \begin{minipage}{0.48\linewidth}
    \centering
    \includegraphics[width=\linewidth]{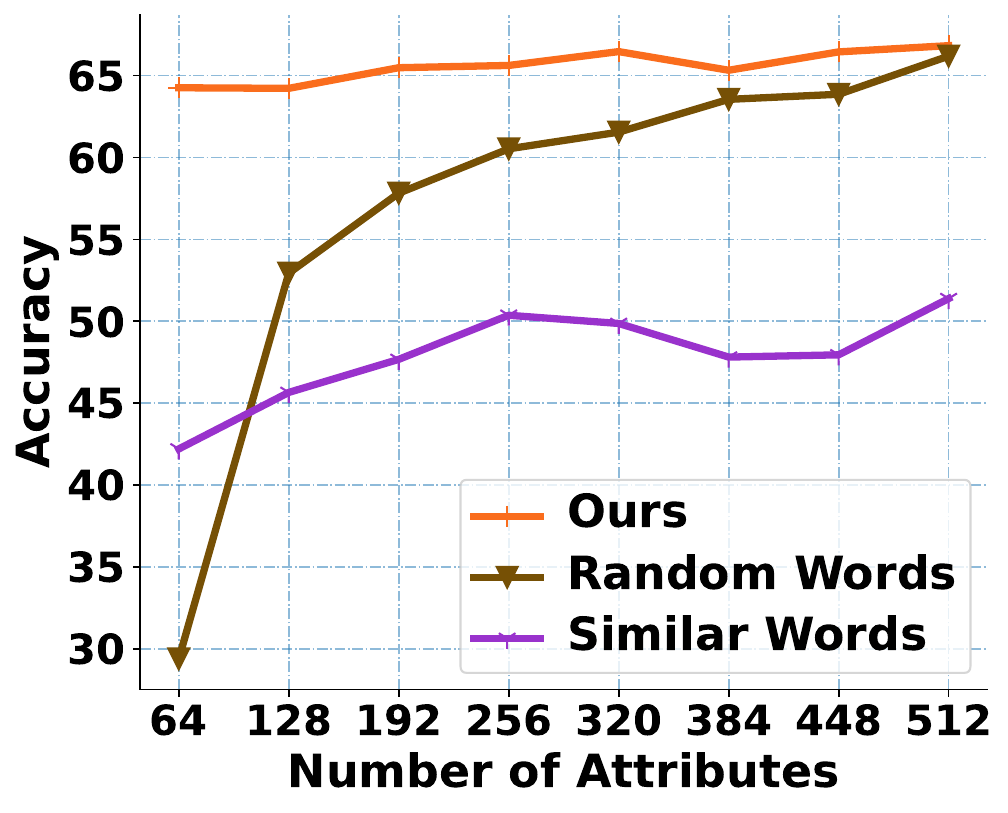}
    \caption{Performance comparison with random or similar words on CUB.}
    \label{fig:compare_gpt3_random_similar}
    \end{minipage}%
    \hfill
    \begin{minipage}{0.47\linewidth}
    \centering
    \captionsetup{type=table} %% tell latex to change to table
    \resizebox{1.0\linewidth}{!}{%
  \begin{tabular}{c|c} 
  \toprule
   & Examples \\
  \midrule
  \multirow{3}{*}{R} & \texttt{boy champagne} \\
  % & {\footnotesize \texttt{condos journal information lexington designs}} \\
  & \texttt{allied whose acrobat}\\
  % & \\
  & {\footnotesize \texttt{eight centered lobby heads} }\\
  \midrule
  \multirow{3}{*}{S} & \texttt{red,gray,snow wings} \\
  & \texttt{orange wings} \\
  & \texttt{lime,navy wings} \\
  \midrule
  \multirow{3}{*}{G} & \texttt{sloping forehead}\\  
  & \texttt{distinctive white throat} \\
  & \texttt{bright red head and breast} \\
  \bottomrule
  \end{tabular}}
    \caption{Examples from Random (R), Silimlar (S), GPT-3 (G) attributes}\label{tab:attribute_instances_from_different_pools}
    \end{minipage}
\end{figure}

\begin{figure*}[t]
\centering
\subfigure[CUB]{\label{fig:cub}\includegraphics[width=0.240\linewidth]{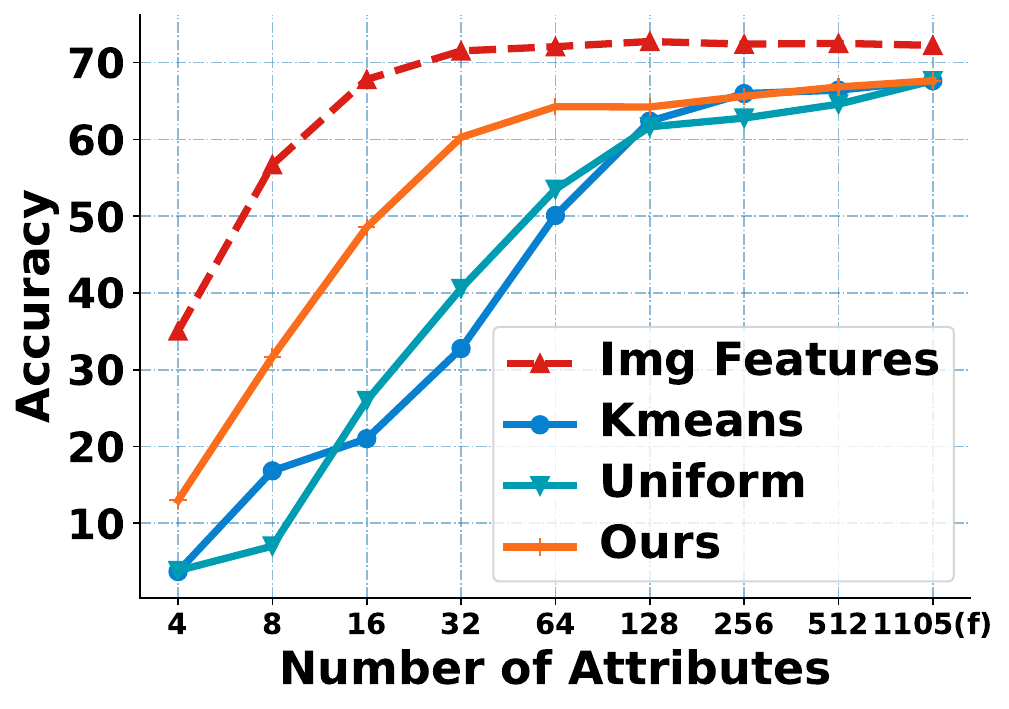}}
\hfill
\subfigure[CIFAR10]{\label{fig:cifar10}\includegraphics[width=0.240\linewidth]{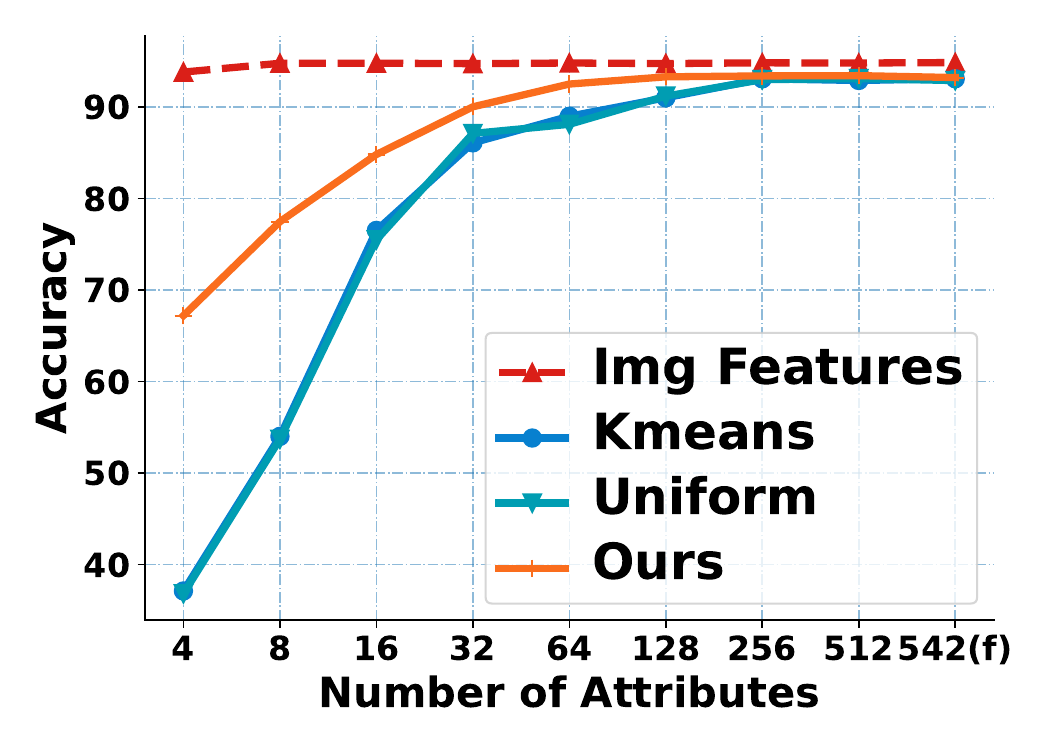}}
\subfigure[CIFAR-100]{\label{fig:CIFAR-100}\includegraphics[width=0.240\linewidth]{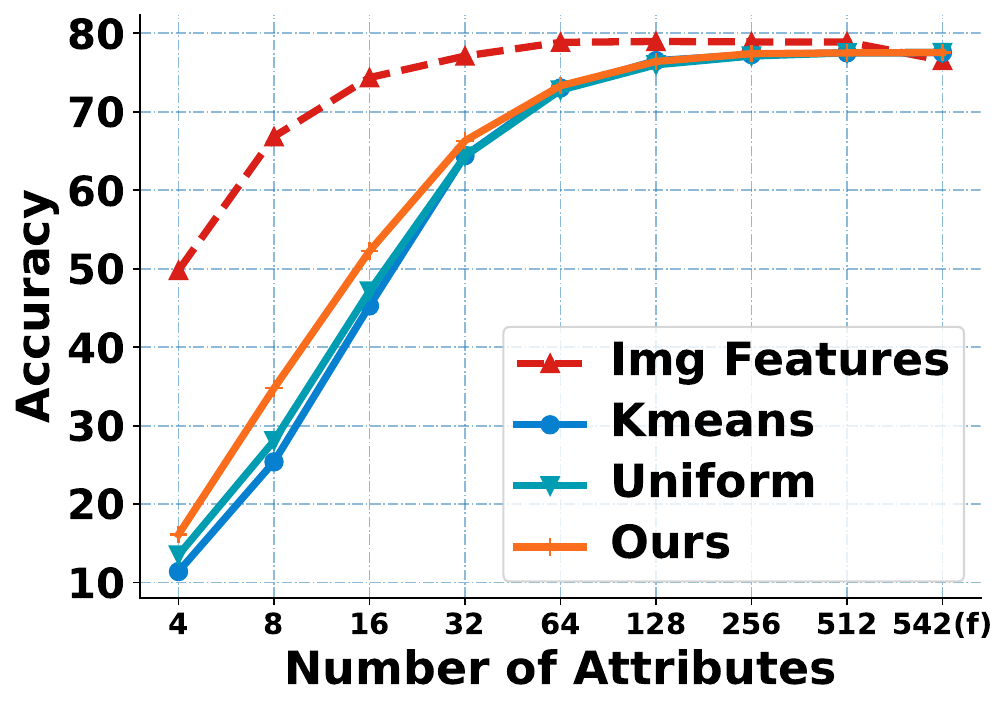}}
\hfill
\subfigure[Flower]{\label{fig:flower}\includegraphics[width=0.240\linewidth]{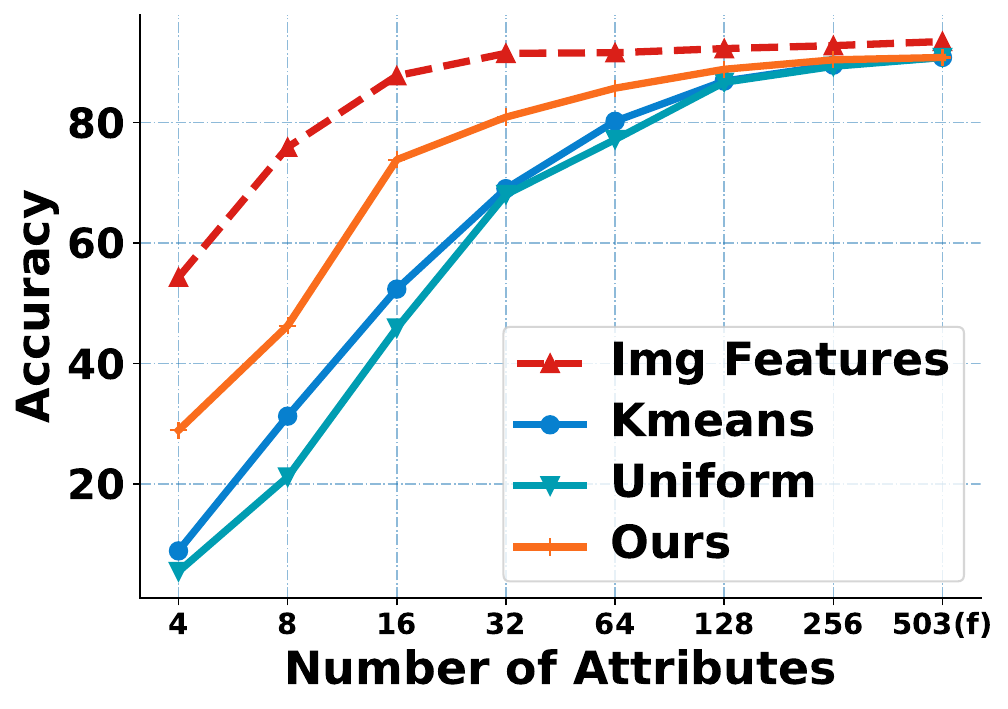}}
\hfill
\subfigure[Food]{\label{fig:food}\includegraphics[width=0.240\linewidth]{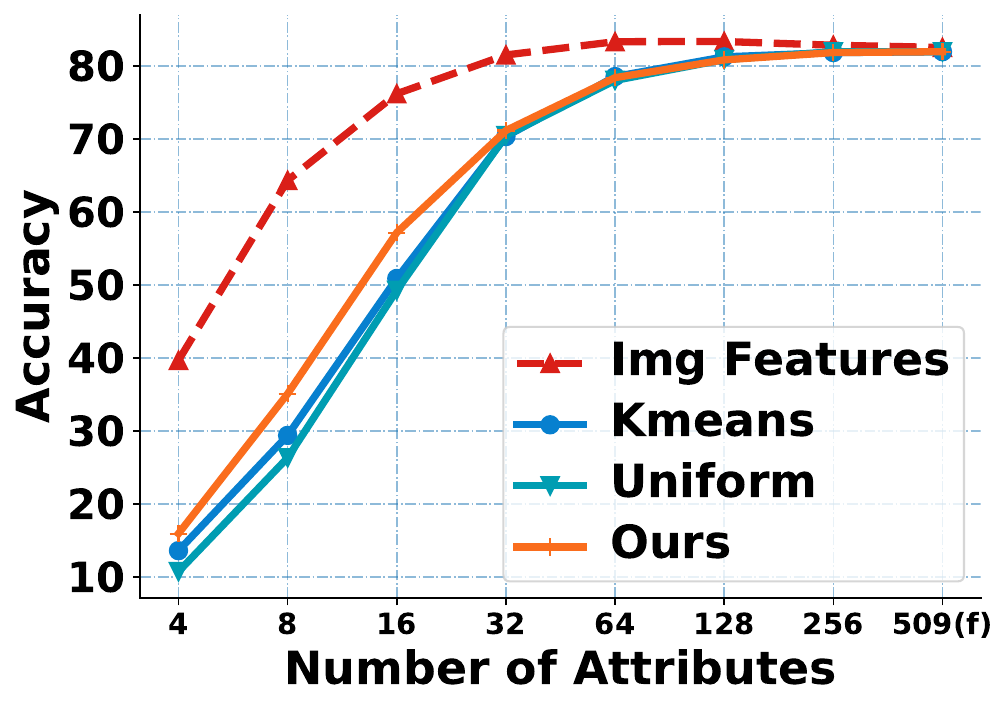}}
\subfigure[Oxford\_pets]{\label{fig:oxford_pets}\includegraphics[width=0.240\linewidth]{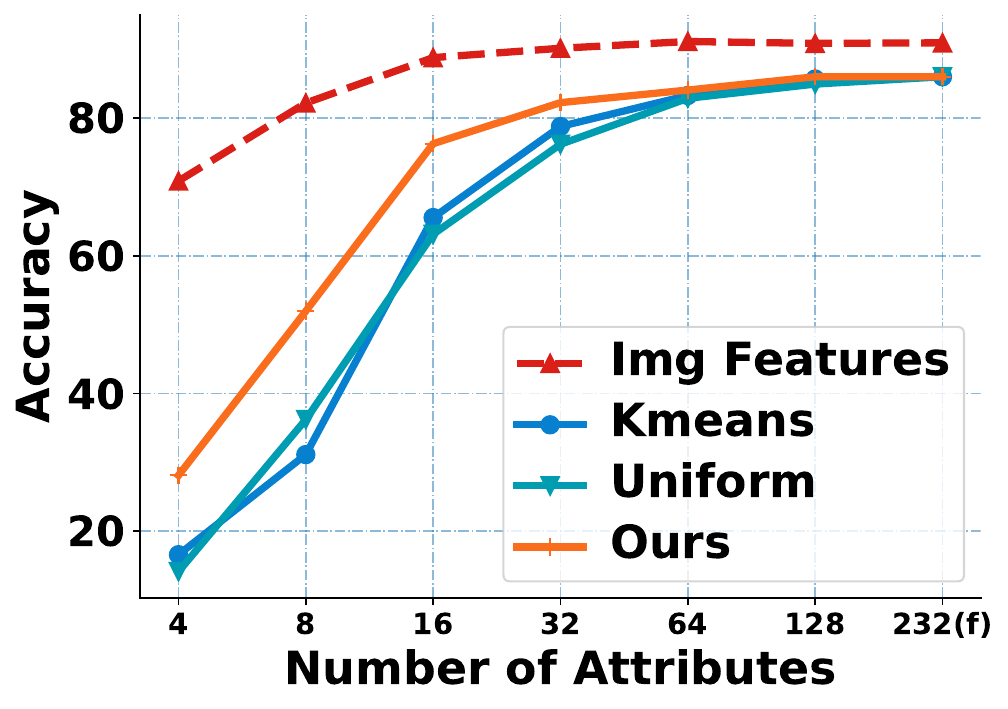}}
\subfigure[Stanford\_cars]{\label{fig:stanford_cars}\includegraphics[width=0.240\linewidth]{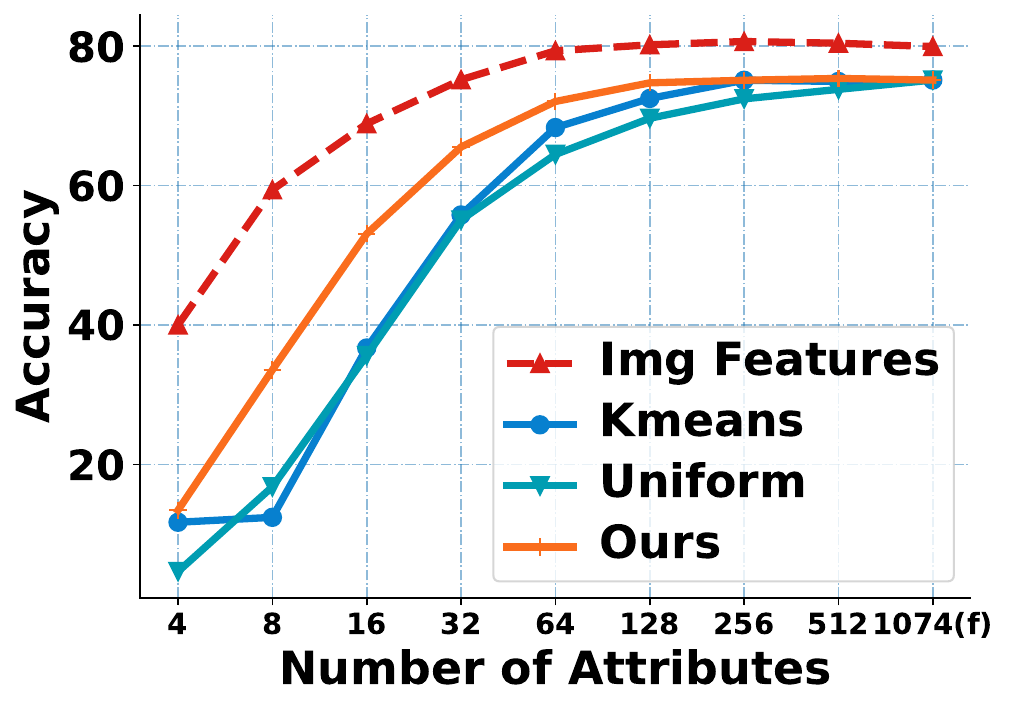}}
\subfigure[Imagenet\_Animals]{\label{fig:imagenet_animals}\includegraphics[width=0.240\linewidth]{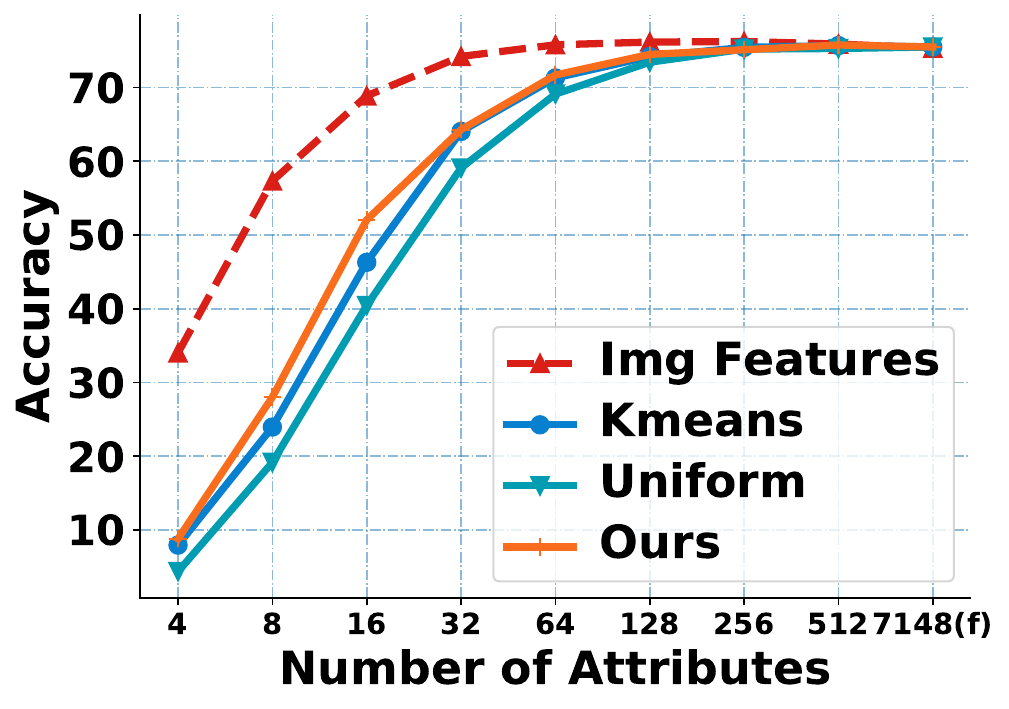}}
\caption{Overall Performance on all datasets. X-axis: number of attributes, Y-axis: Accuracy (\%), ``(f)" means ``full", \ie, all attributes in the pool are used. Uniform refers to uniform sampling. }
\label{fig:overall_performances}
\end{figure*}

We present our finding that LLM-generated attributes in a large quantity behave like random words. Specifically, we compare our method of using GPT-3 attributes with random or similar words. Here, we constructed random words by randomly choosing 1-5 words from the entire English vocabulary, and semantically similar words by combining 1-3 random colors with the noun ``wings'' as suffix. As shown in~\cref{fig:compare_gpt3_random_similar}, when $K=512$, random words perform as well as GPT-3 attributes in terms of classification accuracy. 
Even reducing $K$ from 512 to 256 does not significantly hurt its performance. 
But when $K$ is small (e.g., 64), the performance of random words drops dramatically.
We conjecture that it is because text embeddings randomly drawn from CLIP are nearly orthogonal bases~\cite{wang2021average}. Given an image feature $\in \mathbb{R}^D$, projection with a set of $K$=$D$ orthogonal bases can perfectly preserve its information. 
We further explore how similar words (e.g., red wings, yellow wings) behave. Embeddings of similar words in a trained language model are not orthogonal bases hence the projection will lose information when $K$ is large (e.g., intuitively it is hard to classify 200 bird species using only the color combination of wings). But as $K$ gets smaller, since those similar words have close semantic meanings, they start to outperform random words.
Overall, these findings motivate us to find a concise set of meaningful attributes while maintaining competitive performance.

\vspace{-5pt}
\paragraph{Number of attributes and selection methods}
Finally, we study performance change under different number of attributes in~\cref{fig:overall_performances}. 
First, our method is competitive with image features when $K$ is large. Reducing number of attributes $K$ to the number of classes $C$ (e.g., 512 to 128 for CUB) does not significantly hurt performance, even for baseline methods. This validates our hypothesis that there is plenty of redundant information in the semantic space when the number of attributes is large (as used in LaBO~\cite{yang2022language}). It is possible to find a subset of expressive attributes for visual recognition. 
Second, we also consistently outperform other methods such as K-means clustering and uniform sampling, demonstrating the effectiveness of our task-guided searching method. 
Third, a heuristic design such as K-means performs similar as uniform selection.
Note that though there is a performance gap between image features and using attributes, the gap can be minimized by using a stronger VLM, as the classification accuracy of attributes relies on the accurate estimation of the correlation between images and attributes ( see more results in~\cref{app:ablation} ).
% there is still a gap between our method and  latent image features, which is a reasonable trade-off between interpretability and classification performance but indicates potential for improvement in the future.

\subsection{Ablation Study}
\label{sec:ablation}

% discriminativeness. 

% As the attributes generated by LLMs performs similar as the random words, this suggests that though these attributes are semantically meaningful, they are also diverse to the extent of random words. 

\paragraph{Robustness to the attribute pool}
First, we aim to explore the effects of different initialized attribute concept pools generated by LLMs. On CUB and CIFAR-100, we compare two attribute pools, attributes generated from classes in each dataset, and attributes generated from the full set of ImageNet classes. As shown in~\cref{tab:gpt3_imagenet}, even with the large and noisy attributes from ImageNet, our method can still efficiently find a small number of representative attributes for a task, and obtains competitive classification performance.

\begin{table}[]
    \centering
    \resizebox{\linewidth}{!}{%
    \begin{tabular}{c|ccc|cccc}
        \toprule
        Datasets & \multicolumn{3}{|c|}{CUB} & \multicolumn{3}{c}{CIFAR-100} \\ 
        \midrule
        K & 8 & 16 & 32 & 8 & 16 & 32 \\
        \midrule
        GPT-3 & \textbf{31.67} & 48.55 & 60.27 & \textbf{34.77} & \textbf{52.24} & \textbf{66.30} \\
        GPT-3-Imagenet & 30.81 & \textbf{49.29} & \textbf{60.41} & 33.80 & 51.01 & 65.61 \\
%         ChatGPT & 21.66 & 40.28 & 47.46 & 
% 33.79 & 51.26 & \textbf{67.06} \\
        \bottomrule
    \end{tabular}}
    \caption{Ablation study \emph{w.r.t.} different concept pools.}
    % \vspace{-8pt}
    \label{tab:gpt3_imagenet}
\end{table}

\begin{figure*}[t]
\subfigure[Least Auklet]
{\label{fig:least_auklet}\includegraphics[width=0.475\linewidth]{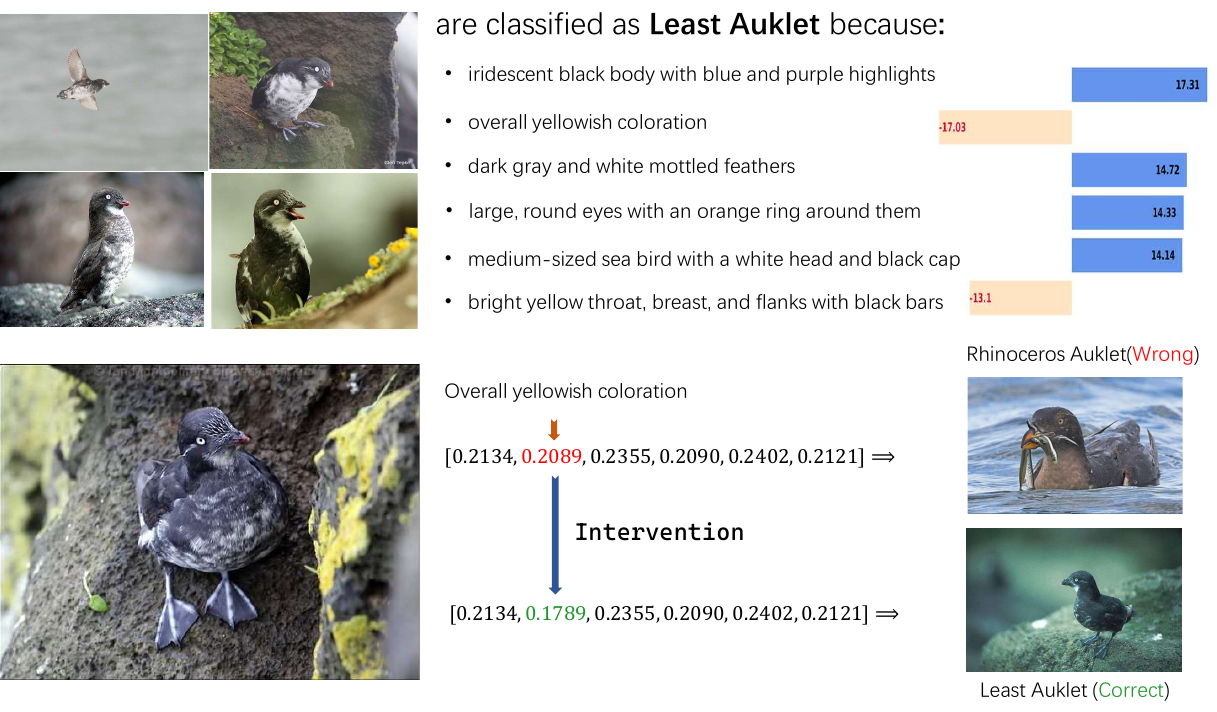}}
\subfigure[Yellow Billed Cuckoo]{\label{fig:yellow_billed_cuckoo}\includegraphics[width=0.475\linewidth]{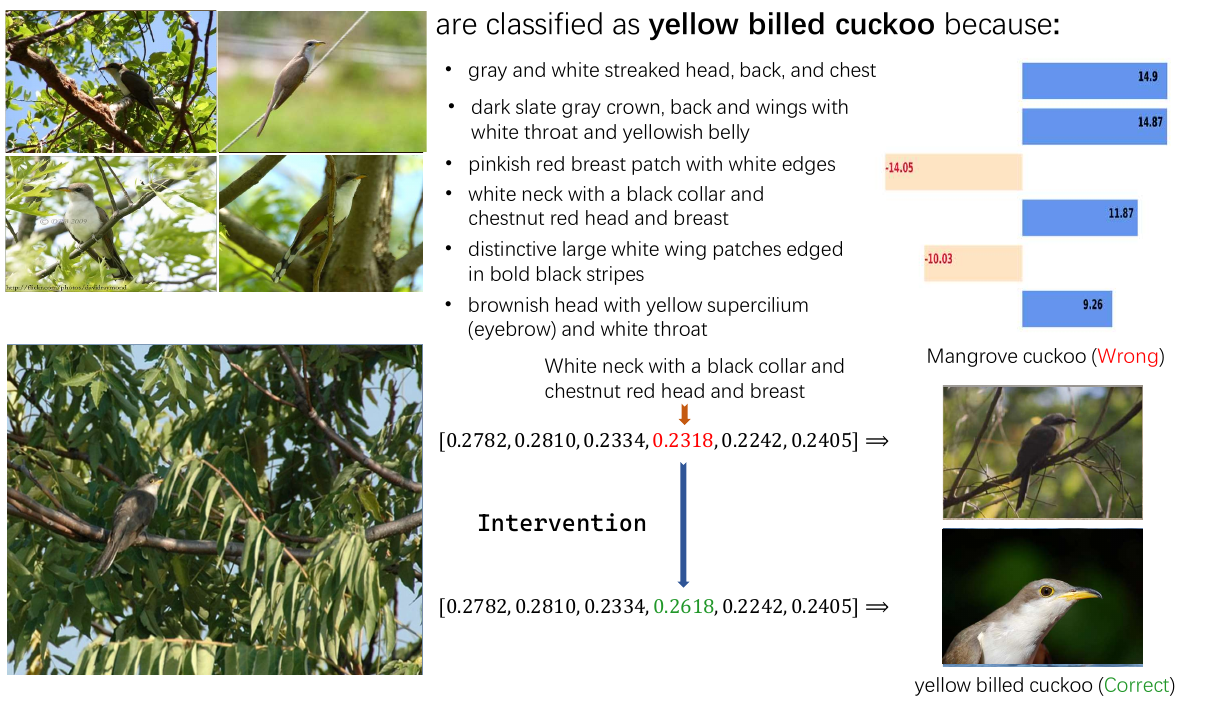}}
\caption{Examples on interpretability and interactivity. (1)~The upper half of each figure show important attributes for two classes of birds. We choose 6 out of 32 attributes with highest importance scores, which are computed by multiplication between clip scores and weights in the linear probe, defined in~\cref{eqn:importance_score}. 
% while the scores are the element-wise multiplication of the semantic vector (consists of clip similarity scores) and the class-related row in the weight of the FC layer in linear probing. 
(2)~The lower half of each figure demonstrates the intervention on the semantic vector (i.e., CLIP scores) to correct the prediction, we use $\delta$=0.03 for all interventions on clip scores as an empirical value. The array of 6 scores are of the same order as the attributes.}
% \vspace{-8pt}
\label{fig:test_time_intervene}
\end{figure*}

% \vspace{-5pt}
\paragraph{Effectiveness of learning-to-search}
Then, we discuss possible choices for selection out of the large attribute pool. Results are shown in~\cref{tab:variation_of_attribute_selection} with the following observations: heuristic methods such as K-means and SVD are not optimal choices for identifying the most distinctive attributes. In fact, they are sometimes less effective than uniform sampling. This is likely because we need to identify the most distinguishing attributes for visual recognition, rather than the most diverse ones based on text embeddings. Overall, our method significantly outperforms other baseline selection methods, showing its efficacy. 
% Our method proves to be superior to the "Similarity" baseline, indicating that relying solely on the average relevance of attributes may not yield optimal results. Intuitively, the most distinct attributes are better suited for image classification, and they are not necessarily the most relevant. (3) 

\begin{table}[]
    \centering
    \resizebox{\linewidth}{!}{%
    \begin{tabular}{c|ccc|cccc}
        \toprule
        Datasets & \multicolumn{3}{|c|}{CUB} & \multicolumn{3}{c}{CIFAR-100} \\ 
        \midrule
        K & 8 & 16 & 32 & 8 & 16 & 32 \\
        \midrule
        K-means & 16.83 & 21.02 & 32.76 & 25.39 & 45.26 & 64.41 \\
        Uniform & 7.02 & 25.98 & 40.58 & 28.07 & 47.14 & 64.34 \\
        SVD & 6.52 & 20.02 & 35.83 & 29.06 & 50.00 & 64.99\\
        Similarity & 4.73 & 9.72 & 18.00 &26.75 & 45.61 & 62.79 \\
        Ours & \textbf{31.67} & \textbf{48.55} & \textbf{60.27} & \textbf{34.77} & \textbf{52.24} & \textbf{66.30} \\
        \bottomrule
    \end{tabular}}
    \caption{Ablation study \emph{w.r.t.} different attribute selection strategies.}
    % \vspace{-4pt}
    \label{tab:variation_of_attribute_selection}
\end{table}
% \vspace{-8pt}

\vspace{-10pt}
\paragraph{Effectiveness of regularization}
% Yu
We compare the Mahalanobis distance (\textit{MAH}) with two variations: (1) \textit{COS}: For each vector $\textbf{E}_j$ and $\textbf{T}_i$ (of the $i$-th attribute) in the concept pool, we computed averaged cosine distance as follows:
\begin{align*}
    \mathcal{L}_{cos} &= \frac{1}{K^2}\sum_{j=1}^K \sum_{i=1}^K \frac{\Trans{\textbf{T}}_i \textbf{E}_j}{||\textbf{T}_i|| || \textbf{E}_j||}
\end{align*}
(2) \textit{CE}: Learning with~\cref{eq:loss_cc} only.
Results are in~\cref{tab:variation_of_regularization}. Overall, Mahalanobis distance is an effective constraint to encourage the dictionary $E$ to be close to the distribution of CLIP embeddings.
\begin{table}[]
    \centering
    \resizebox{0.65\linewidth}{!}
    {%
    \begin{tabular}{c|cccc}
        \toprule
        Dataset & \multicolumn{4}{c}{CUB} \\ 
        \midrule
        K & 8 & 16 & 32 & 64\\
        \midrule
        \textit{MAH} & 30.76 & 47.87 & \textbf{60.27} & \textbf{64.25} \\
        \textit{COS} & 28.96 & 47.35 & 58.27 & 63.25\\
        % $\mathcal{L}_2$ & 28.96 & 47.35 & 55.68 & 60.39 & 31.98 & 51.15 & 65.02 &  72.80 \\
        \textit{CE} & \textbf{31.67} & \textbf{48.55} & 55.88 & 60.73\\
        \midrule
        Dataset & \multicolumn{4}{c}{CIFAR-100} \\ 
        \midrule
        K & 8 & 16 & 32 & 64\\
        \midrule
        \textit{MAH} & \textbf{34.77} & \textbf{52.24} & 65.91 & \textbf{73.31} \\
        \textit{COS} & 31.98 & 51.15 & 65.02 &  72.80\\
        % $\mathcal{L}_2$ & 28.96 & 47.35 & 55.68 & 60.39 & 31.98 & 51.15 & 65.02 &  72.80 \\
        \textit{CE} & 32.45 & 50.83 & \textbf{66.29} & 73.25 \\
        \bottomrule
    \end{tabular}}
    \caption{Ablation study \emph{w.r.t.} different regularization.}
    \label{tab:variation_of_regularization}
\end{table}
% \vspace{-5pt}

% \paragraph{Comparing with zero-shot learning}
% \begin{table}[h!]
% \vspace{-10pt}
%     \centering
%     \resizebox{\linewidth}{!}{%
%     \begin{tabular}{c cccc}   
%         \toprule
%         & CIFAR-100 & Stanford-cars & Flower & Imagenet-Ani \\ 
%         \midrule
%         CLIP-ZS w/ class names & 54.49 & 57.87 & 60.19 & 59.44\\
%         CLIP-ZS w/ attributes  & 30.07 & 5.42 & 9.80 & 9.13\\
%          \midrule
%          CLIP-Train Visual  & 79.30 & 79.95 & 92.35 & 75.31 \\
%          Ours (K=512) & 75.41 & 74.67 & 90.29 & 75.60 \\
%          \bottomrule
%     \end{tabular}}
%     \vspace{-10pt}
%     \caption{Comparison with zero-shot classification methods.}
%    \label{tab:comparison_with_zero_shot_clip}
% \end{table}

% We deliver more results in Table~\ref{tab:comparison_with_zero_shot_clip}. We use \textit{A photo of} as the prompt for all methods. Zero-shot (CLIP-ZS) is worse than supervised training. Note that CLIP-ZS with class names may not be a fair comparison, \textbf{as our goal is to classify images with attributes instead of class names}, thereby gaining a level of interpretability and fine-grained understanding of visual recognition. If we use only attributes for CLIP-ZS, the performance drastically decreases.

\subsection{Analysis of Interpretability and Interactivity}
\noindent We perform analysis and visualizations to show that: \\
(1) \textbf{Our learned attributes provide interpretability}. 
As shown in~\cref{fig:test_time_intervene}, the upper half presents the images in a class $c$ and high relevant attributes to recognize them. 
Specifically, we denote $\textbf{W} \in \mathbb{R}^{K_C*K}$ as the weight of the FC layer in linear probing, where $K_C$, $K$ are the number of classes and attributes. 
Then for each image $i$ and its semantic vector $\textbf{A} \in \mathbb{R}^K$, 
we multiply the corresponding score vector of image $i$ with the corresponding row of the FC layer $\textbf{W}_c$ to compute Importance Score $\textbf{IS} \in \mathbb{R}^{K}$:
\begin{equation}
    \textbf{IS} = \textbf{W}_c \otimes \textbf{A}
\label{eqn:importance_score}
\end{equation}
where $\otimes$ means element-wise multiplication. Then we present attributes with the top absolute values of $\textbf{IS}$ averaged over all samples in a class from the test set, with blue/orange bars indicating the positive/negative importance. 
% Since all CLIP scores~\cite{hessel2021clipscore} are positive, our linear probe also gains interpretability, as classes with larger logits indicate higher probabilities for prediction.
% Note that higher absolute values mean higher importance, while positive ones means this attribute is important to the class, while negative ones indicate that this image should not be classified as this class if it's highly aligned with the current attribute.
Higher absolute values denote greater significance. Since all CLIP scores are positive~\cite{hessel2021clipscore}, the positivity or negativity of high IS signifies their relevance to the class.

\begin{figure*}
    \centering
    \includegraphics[width=\linewidth]{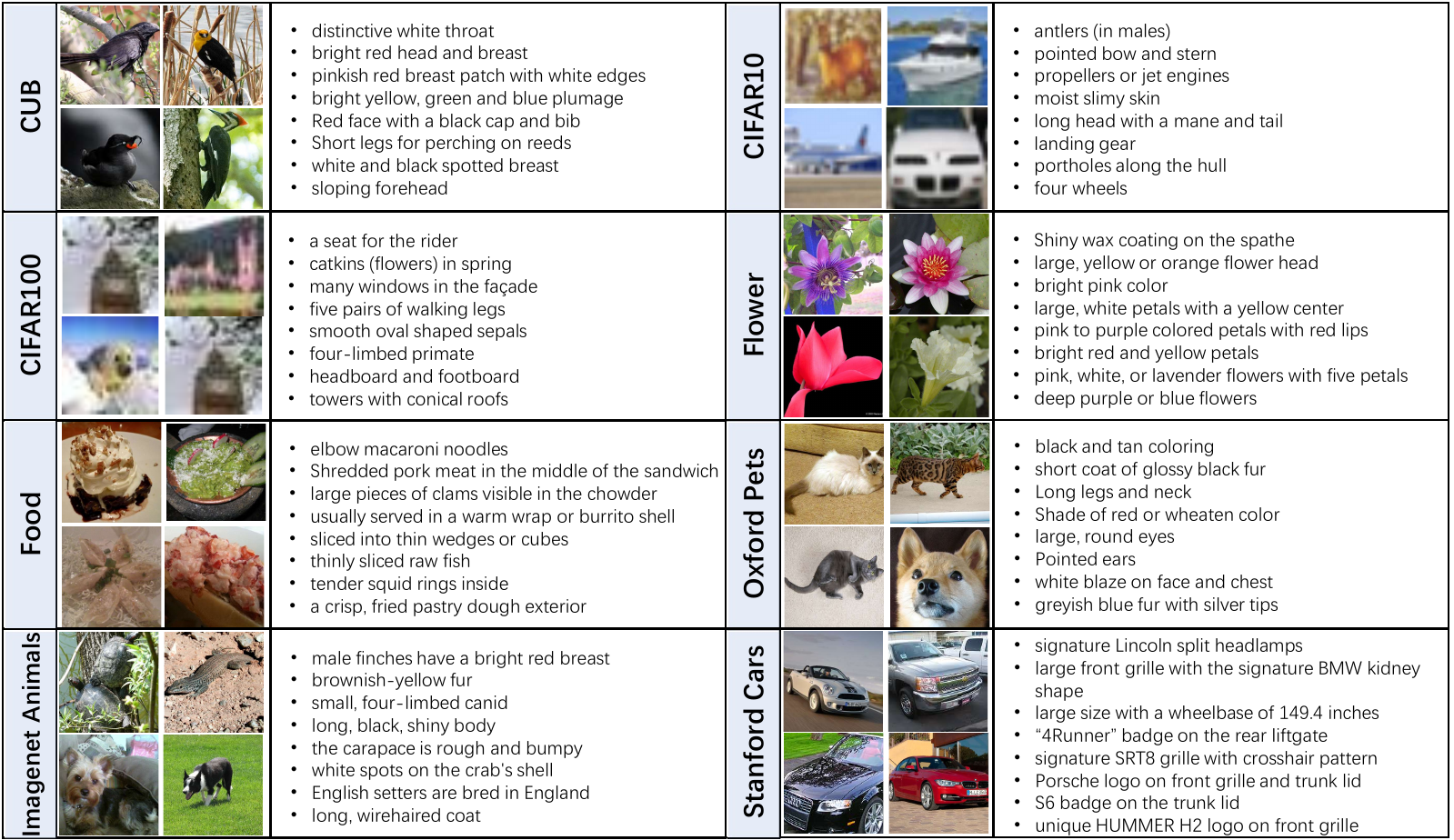}
    \caption{A concise set of 8 descriptive attributes learned for each dataset with sampled images.}
    \label{fig:pruned_features}
\end{figure*}

\noindent (2) \textbf{Our concise set of attributes enables simple interactivity}. As shown in the lower half of~\cref{fig:test_time_intervene}, we can correct the model's wrong predictions during inference by changing only a single similarity score between an image and the attribute that the CLIP model made a mistake on. This is a significant simplification compared with previous work~\cite{koh2020concept} where they need to manipulate scores from a group of concepts for the CUB dataset. We present more user studies in~\cref{app:case_study}.

% This shows that human could interact with the model by intervening on the similarity scores to help the model make right predictions. 
% We present more case studies in Appendix \ref{app:case_study}

% Manipulating the concept scores during test time: Say an image with bird-A is classified as bird-B by our model, by changing some similarity scores in the vector, we can correct the model's prediction back to bird-A. 

% \begin{figure}
%     \centering
%     \includegraphics[width=\linewidth]{figures/least_auklet.pdf}
%     \caption{Important attributes for Least Auklet out of 32 attributes and the intervention on the clip score in the dataset.}
%     \label{fig:least_auklet}
% \end{figure}
% \begin{figure}
%     \centering
%     \includegraphics[width=\linewidth]{figures/yellow_billed_cuckoo.pdf}
%     \caption{Important attributes for Yellow Billed Cuckoo out of 32 attributes and the intervention on the clip score in the dataset.}
%     \label{fig:yellow_billed_cuckoo}
% \end{figure}

\subsection{Visualization of Our Discovered Attributes}
We show our learned descriptive attributes with $K=8$ in~\cref{fig:pruned_features}. Intuitively, we can observe these attributes are distinctive for each domain. Take birds recognition (CUB) as an example, the eight attributes covered most of the body parts of a bird (head, breast, legs, etc.). As we are condensing knowledge from hundreds of bird classes, each attribute broadly covers many categories. A bright red head and breast can be a noticeable visual attribute for many bird species, such as the Northern Cardinal and the Vermilion Flycatcher. Overall, explaining a domain with a few descriptive attributes is challenging, even for an expert with sufficient domain knowledge. But our model is able to automatically provide a level of knowledge to help humans understand how visual recognition works. 

\begin{figure}
    \centering
    \includegraphics[width=\linewidth]{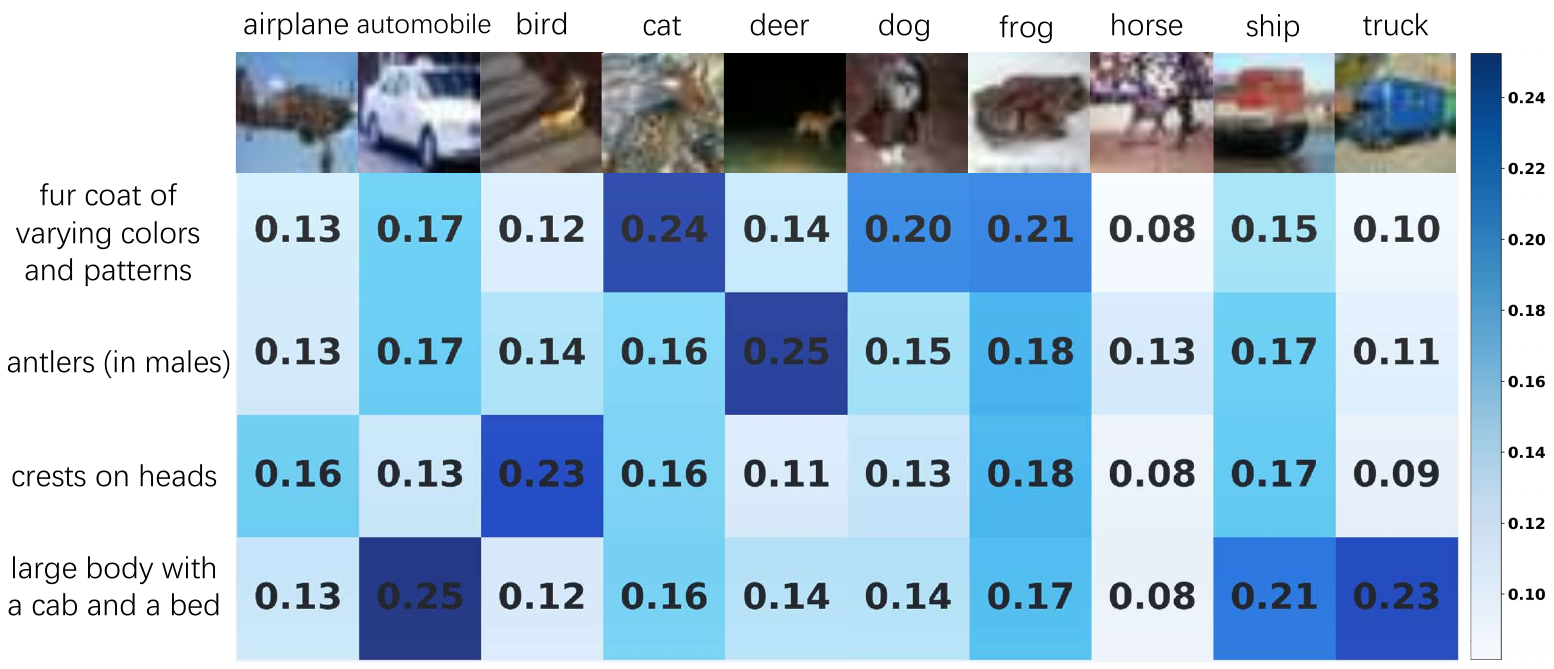}
    \vspace{-8pt}
    \caption{Case study on CIFAR-10. The numbers are CLIP similarity scores between each image and attributes.}
    \label{fig:cifar10_case_study} 
    % \vspace{-8pt}
\end{figure}

We then present case studies on CIFAR-10 with 4 attributes and CLIP scores of 10 random images from each class in~\cref{fig:cifar10_case_study}.
In general, each image is activated in an distinguishable way in the heat map. Some attributes can distinguish a few classes, for example, cat and dog have higher activation on ``fur coat'' compared to automobile or truck. Thus ``fur coat'' may be an important feature to differentiate animals and vehicles.
% These results suggest that our method can effectively identify the descriptive attributes that distinguish different types of objects.

% \paragraph{Similarity scores under adversarial attack / adding noises}
\section{Related work}
\paragraph{Interpretable Deep Learning}
Interpretability is a critical research problem for deep learning with black-box models~\cite{DBLP:journals/corr/abs-2105-05328,phillips2020four,LIME,Anchors,DBLP:journals/corr/abs-1805-10820,Barnett_nature,PA3D}. 
% Among them, text descriptions are shown to have the potential to provide better interpretability than visual explanations~\cite{selvaraju2016grad,selvaraju2017grad}, by providing explanations via natural language.
Some works study model behavior and explore if deep models could encode concepts for understanding~\cite{TCAV,IJCNN_CAV,L2C,ACKinAlphaZero}. 
For image classification, preliminary attempts aim to describe objects with attributes~\cite{farhadi2009describing,Lampert_2009,Kumar_2009} or building concept bottleneck models~\cite{koh2020concept, compDL,PCBM,Concept_whitening}. These methods require in-depth human analysis and intensive labeling, which are impractical to scale to more classes and domains. 
% to partially alleviate the black-box nature of machine learneural models. 

Recent works~\cite{menon2022visual,pratt2022does,yang2022language} tackle this problem by using GPT-3 as a knowledge base to query visual attributes or concepts. Specifically,~\cite{menon2022visual,pratt2022does} generate descriptions with LLMs, and use them for knowledge-aware prompting for each class to improve zero-shot performance of CLIP~\cite{radford2021learning}. For example, given the class name ``bee'', it will augment it with attributes such as ``A bee with black and yellow body''. Our work differs in that our goal is to learn representative attributes for visual recognition without using class names. 
LABO~\cite{yang2022language} extends the idea of concept bottleneck models by generating thousands of concepts from LLMs.
% However, the large space of GPT-3 descriptions is hard for humans to understand and interact with.
% Moreover, we found the semantic vectors contain redundant information in high dimensional spaces, and we are able to use only a small number of attributes to achieve similar classification performance.
Inspired by our finding that there is great redundancy in the large-scale attributes, we aim to learn a concise set of attributes that are initially generated from LLMs for each task, while maintaining the classification performance as possible. Concise attributes also enable stronger interpretability and interactivity, and can help humans to summarize critical knowledge for visual recognition in an automatic way.

\paragraph{Foundation Models}
Recently, foundation models~\cite{bommasani2021opportunities}, which are pre-trained with a large amount of data and large model sizes, have revolutionalized machine learning research and many fields. These models are shown to be adaptable to a wide range of downstream tasks for computer vision~\cite{he2022masked,wei2022masked,zhou2021ibot}, natural language processing~\cite{devlin2018bert,chung2022scaling,zhang2022opt,yan2022radbert} and cross-modal research~\cite{li2022blip,LXMERT,ALIGN,guzhov2022audioclip}. 
One direction is to train LLMs such as GPT3~\cite{brown2020language} and ChatGPT with massive text to serve as a powerful knowledge base with high interactivity and beyond.
Another direction is to build VLMs~\cite{radford2021learning,UniCL,florence,CoCa,alayrac2022flamingo}, which connect vision and language by pre-training with image-text pairs and learning a joint embedding space for both. % There are generative VLMs~\cite{prefixcondition,CoCa,Uniter,git,simvlm,VLP,alayrac2022flamingo} but they are not trained to provide world knowledge but mainly vision language alignment. 
In this work, we use LLMs as a knowledge base for querying visual related knowledge, and use VLMs to bridge vision and text, presenting a new paradigm for interpretable visual recognition in the era of foundation models.
% In our paper, we use LLMs as a knowledge base for querying visual related knowledge and use VLMs to bridge vision and text for visual recognition. 

\section{Discussion}
There are many interesting topics to explore with our new paradigm.
First, our framework is a plug-and-play model that can be readily applied to many other vision tasks, by simply changing the task-guided learning objective to a particular task, e.g., classification losses for object detection, video understanding, and 3D classification. Furthermore, a concise set of descriptive attributes enables interactivity for vision models and empowers human-machine cooperation in a user-friendly way through natural language interfaces.
Lastly, we show the potential of summarizing knowledge for challenging vision tasks in the new era of LLMs, which could have broad impact for various domains.

% Our new paradigm has several interesting implications and potential applications. First, it is a plug-and-play model that can be easily adapted to various vision tasks by modifying the learning objective. For example, it can be applied to object detection, video understanding, and 3D classification by changing the classification losses. Additionally, the use of a concise set of descriptive attributes enables the model to be interactive and facilitates human-machine cooperation through natural language interfaces. Finally, our work demonstrates the potential to summarize knowledge for challenging vision tasks in the era of LLMs, which could have broad impact across different domains.

% There are many interesting topics to explore with our new paradigm.
% First, our framework is a plug-and-play model that can be readily applied to many other vision tasks, by simply changing the task-guided learning objective to a particular task, e.g., classification losses for object detection, video understanding, and 3D classification. Furthermore, a concise set of descriptive attributes can empower the current vision model to be interactive and enables human-machine cooperation in a user-friendly way through natural language interfaces.
% Lastly, we show the potential of summarizing knowledge for challenging vision tasks in the new era of LLMs, with the potential to impact various domains.

% much more opportunities for human model interaction comparing with previous work. 
% We can now easily interact with the model through a few key descriptions as a natural language interface. 
\section{Conclusion}
In this work, we propose a new paradigm for visual recognition that leverages a concise set of descriptive attributes. Motivated by our insightful finding that significant redundancy exists in massive LLMs-generated attributes, we design a simple yet effective searching method guided by image-level labels, to identify an informative subset. Our new paradigm is validated across 8 datasets to achieve strong classification accuracy with multiple benefits and broad impacts, including efficiency, interpretability, human interactivity, and knowledge summarization.

\section*{Acknowledgments}
We would like to sincerely thank the anonymous reviewers and chairs for their careful review of our work, with helpful and constructive suggestions to improve the paper.

% \clearpage
{\small
\bibliographystyle{ieee_fullname}
\bibliography{egbib}

\begin{thebibliography}{10}\itemsep=-1pt

\bibitem{alayrac2022flamingo}
Jean-Baptiste Alayrac, Jeff Donahue, Pauline Luc, Antoine Miech, Iain Barr,
  Yana Hasson, Karel Lenc, Arthur Mensch, Katie Millican, Malcolm Reynolds,
  et~al.
\newblock Flamingo: a visual language model for few-shot learning.
\newblock {\em arXiv preprint arXiv:2204.14198}, 2022.

\bibitem{Barnett_nature}
Alina~Jade Barnett, Fides~Regina Schwartz, Chaofan Tao, Chaofan Chen, Yinhao
  Ren, Joseph~Y. Lo, and Cynthia Rudin.
\newblock A case-based interpretable deep learning model for classification of
  mass lesions in digital mammography.
\newblock {\em Nat. Mach. Intell.}, 3(12):1061--1070, 2021.

\bibitem{bommasani2021opportunities}
Rishi Bommasani, Drew~A Hudson, Ehsan Adeli, Russ Altman, Simran Arora, Sydney
  von Arx, Michael~S Bernstein, Jeannette Bohg, Antoine Bosselut, Emma
  Brunskill, et~al.
\newblock On the opportunities and risks of foundation models.
\newblock {\em arXiv preprint arXiv:2108.07258}, 2021.

\bibitem{food}
Lukas Bossard, Matthieu Guillaumin, and Luc~Van Gool.
\newblock Food-101 - mining discriminative components with random forests.
\newblock In {\em {ECCV} {(6)}}, volume 8694 of {\em Lecture Notes in Computer
  Science}, pages 446--461. Springer, 2014.

\bibitem{brown2020language}
Tom Brown, Benjamin Mann, Nick Ryder, Melanie Subbiah, Jared~D Kaplan, Prafulla
  Dhariwal, Arvind Neelakantan, Pranav Shyam, Girish Sastry, Amanda Askell,
  et~al.
\newblock Language models are few-shot learners.
\newblock {\em Advances in neural information processing systems},
  33:1877--1901, 2020.

\bibitem{Concept_whitening}
Zhi Chen, Yijie Bei, and Cynthia Rudin.
\newblock Concept whitening for interpretable image recognition.
\newblock {\em Nat. Mach. Intell.}, 2(12):772--782, 2020.

\bibitem{chung2022scaling}
Hyung~Won Chung, Le Hou, Shayne Longpre, Barret Zoph, Yi Tay, William Fedus,
  Eric Li, Xuezhi Wang, Mostafa Dehghani, Siddhartha Brahma, et~al.
\newblock Scaling instruction-finetuned language models.
\newblock {\em arXiv preprint arXiv:2210.11416}, 2022.

\bibitem{cimpoi2014describing}
Mircea Cimpoi, Subhransu Maji, Iasonas Kokkinos, Sammy Mohamed, and Andrea
  Vedaldi.
\newblock Describing textures in the wild.
\newblock In {\em Proceedings of the IEEE conference on computer vision and
  pattern recognition}, pages 3606--3613, 2014.

\bibitem{imagenet}
Jia Deng, Wei Dong, Richard Socher, Li{-}Jia Li, Kai Li, and Li Fei{-}Fei.
\newblock Imagenet: {A} large-scale hierarchical image database.
\newblock In {\em {CVPR}}, pages 248--255. {IEEE} Computer Society, 2009.

\bibitem{devlin2018bert}
Jacob Devlin, Ming-Wei Chang, Kenton Lee, and Kristina Toutanova.
\newblock Bert: Pre-training of deep bidirectional transformers for language
  understanding.
\newblock {\em arXiv preprint arXiv:1810.04805}, 2018.

\bibitem{DBLP:journals/corr/abs-2105-05328}
Jack Dunn, Luca Mingardi, and Ying~Daisy Zhuo.
\newblock Comparing interpretability and explainability for feature selection.
\newblock {\em CoRR}, abs/2105.05328, 2021.

\bibitem{farhadi2009describing}
Ali Farhadi, Ian Endres, Derek Hoiem, and David Forsyth.
\newblock Describing objects by their attributes.
\newblock In {\em 2009 IEEE conference on computer vision and pattern
  recognition}, pages 1778--1785. IEEE, 2009.

\bibitem{DBLP:journals/corr/abs-1805-10820}
Riccardo Guidotti, Anna Monreale, Salvatore Ruggieri, Dino Pedreschi, Franco
  Turini, and Fosca Giannotti.
\newblock Local rule-based explanations of black box decision systems.
\newblock {\em CoRR}, abs/1805.10820, 2018.

\bibitem{guzhov2022audioclip}
Andrey Guzhov, Federico Raue, J{\"o}rn Hees, and Andreas Dengel.
\newblock Audioclip: Extending clip to image, text and audio.
\newblock In {\em ICASSP 2022-2022 IEEE International Conference on Acoustics,
  Speech and Signal Processing (ICASSP)}, pages 976--980. IEEE, 2022.

\bibitem{he2022masked}
Kaiming He, Xinlei Chen, Saining Xie, Yanghao Li, Piotr Doll{\'a}r, and Ross
  Girshick.
\newblock Masked autoencoders are scalable vision learners.
\newblock In {\em Proceedings of the IEEE/CVF Conference on Computer Vision and
  Pattern Recognition}, pages 16000--16009, 2022.

\bibitem{hessel2021clipscore}
Jack Hessel, Ari Holtzman, Maxwell Forbes, Ronan~Le Bras, and Yejin Choi.
\newblock Clipscore: A reference-free evaluation metric for image captioning.
\newblock {\em arXiv preprint arXiv:2104.08718}, 2021.

\bibitem{ALIGN}
Chao Jia, Yinfei Yang, Ye Xia, Yi{-}Ting Chen, Zarana Parekh, Hieu Pham,
  Quoc~V. Le, Yun{-}Hsuan Sung, Zhen Li, and Tom Duerig.
\newblock Scaling up visual and vision-language representation learning with
  noisy text supervision.
\newblock In {\em {ICML}}, volume 139 of {\em Proceedings of Machine Learning
  Research}, pages 4904--4916. {PMLR}, 2021.

\bibitem{kim2018interpretability}
Been Kim, Martin Wattenberg, Justin Gilmer, Carrie Cai, James Wexler, Fernanda
  Viegas, et~al.
\newblock Interpretability beyond feature attribution: Quantitative testing
  with concept activation vectors (tcav).
\newblock In {\em International conference on machine learning}, pages
  2668--2677. PMLR, 2018.

\bibitem{TCAV}
Been Kim, Martin Wattenberg, Justin Gilmer, Carrie~J. Cai, James Wexler,
  Fernanda~B. Vi{\'{e}}gas, and Rory Sayres.
\newblock Interpretability beyond feature attribution: Quantitative testing
  with concept activation vectors {(TCAV)}.
\newblock In {\em {ICML}}, volume~80 of {\em Proceedings of Machine Learning
  Research}, pages 2673--2682. {PMLR}, 2018.

\bibitem{kingma2014adam}
Diederik~P Kingma and Jimmy Ba.
\newblock Adam: A method for stochastic optimization.
\newblock {\em arXiv preprint arXiv:1412.6980}, 2014.

\bibitem{kocon2023chatgpt}
Jan Koco{\'n}, Igor Cichecki, Oliwier Kaszyca, Mateusz Kochanek, Dominika
  Szyd{\l}o, Joanna Baran, Julita Bielaniewicz, Marcin Gruza, Arkadiusz Janz,
  Kamil Kanclerz, et~al.
\newblock Chatgpt: Jack of all trades, master of none.
\newblock {\em arXiv preprint arXiv:2302.10724}, 2023.

\bibitem{koh2020concept}
Pang~Wei Koh, Thao Nguyen, Yew~Siang Tang, Stephen Mussmann, Emma Pierson, Been
  Kim, and Percy Liang.
\newblock Concept bottleneck models.
\newblock In {\em International Conference on Machine Learning}, pages
  5338--5348. PMLR, 2020.

\bibitem{stanford_cars}
Jonathan Krause, Michael Stark, Jia Deng, and Li Fei-Fei.
\newblock 3d object representations for fine-grained categorization.
\newblock In {\em 4th International IEEE Workshop on 3D Representation and
  Recognition (3dRR-13)}, Sydney, Australia, 2013.

\bibitem{cifar10}
Alex Krizhevsky, Geoffrey Hinton, et~al.
\newblock Learning multiple layers of features from tiny images.
\newblock 2009.

\bibitem{Kumar_2009}
Neeraj Kumar, Alexander~C. Berg, Peter~N. Belhumeur, and Shree~K. Nayar.
\newblock Attribute and simile classifiers for face verification.
\newblock In {\em {ICCV}}, pages 365--372. {IEEE} Computer Society, 2009.

\bibitem{Lampert_2009}
Christoph~H. Lampert, Hannes Nickisch, and Stefan Harmeling.
\newblock Learning to detect unseen object classes by between-class attribute
  transfer.
\newblock In {\em {CVPR}}, pages 951--958. {IEEE} Computer Society, 2009.

\bibitem{li2022blip}
Junnan Li, Dongxu Li, Caiming Xiong, and Steven Hoi.
\newblock Blip: Bootstrapping language-image pre-training for unified
  vision-language understanding and generation.
\newblock In {\em International Conference on Machine Learning}, pages
  12888--12900. PMLR, 2022.

\bibitem{IJCNN_CAV}
Adriano Lucieri, Muhammad~Naseer Bajwa, Stephan~Alexander Braun, Muhammad~Imran
  Malik, Andreas Dengel, and Sheraz Ahmed.
\newblock On interpretability of deep learning based skin lesion classifiers
  using concept activation vectors.
\newblock In {\em {IJCNN}}, pages 1--10. {IEEE}, 2020.

\bibitem{ACKinAlphaZero}
Thomas McGrath, Andrei Kapishnikov, Nenad Tomasev, Adam Pearce, Demis Hassabis,
  Been Kim, Ulrich Paquet, and Vladimir Kramnik.
\newblock Acquisition of chess knowledge in alphazero.
\newblock {\em CoRR}, abs/2111.09259, 2021.

\bibitem{menon2022visual}
Sachit Menon and Carl Vondrick.
\newblock Visual classification via description from large language models.
\newblock {\em arXiv preprint arXiv:2210.07183}, 2022.

\bibitem{flower}
Maria{-}Elena Nilsback and Andrew Zisserman.
\newblock Automated flower classification over a large number of classes.
\newblock In {\em {ICVGIP}}, pages 722--729. {IEEE} Computer Society, 2008.

\bibitem{openai2022chatgpt}
TB OpenAI.
\newblock Chatgpt: Optimizing language models for dialogue.
\newblock {\em OpenAI}, 2022.

\bibitem{oxford_pets}
Omkar~M. Parkhi, Andrea Vedaldi, Andrew Zisserman, and C.~V. Jawahar.
\newblock Cats and dogs.
\newblock In {\em {CVPR}}, pages 3498--3505. {IEEE} Computer Society, 2012.

\bibitem{phillips2020four}
P~Jonathon Phillips, Carina~A Hahn, Peter~C Fontana, David~A Broniatowski, and
  Mark~A Przybocki.
\newblock Four principles of explainable artificial intelligence.
\newblock {\em Gaithersburg, Maryland}, 2020.

\bibitem{pratt2022does}
Sarah Pratt, Rosanne Liu, and Ali Farhadi.
\newblock What does a platypus look like? generating customized prompts for
  zero-shot image classification.
\newblock {\em arXiv preprint arXiv:2209.03320}, 2022.

\bibitem{radford2021learning}
Alec Radford, Jong~Wook Kim, Chris Hallacy, Aditya Ramesh, Gabriel Goh,
  Sandhini Agarwal, Girish Sastry, Amanda Askell, Pamela Mishkin, Jack Clark,
  et~al.
\newblock Learning transferable visual models from natural language
  supervision.
\newblock In {\em International conference on machine learning}, pages
  8748--8763. PMLR, 2021.

\bibitem{LIME}
Marco~T{\'{u}}lio Ribeiro, Sameer Singh, and Carlos Guestrin.
\newblock "why should {I} trust you?": Explaining the predictions of any
  classifier.
\newblock In {\em {KDD}}, pages 1135--1144. {ACM}, 2016.

\bibitem{Anchors}
Marco~T{\'{u}}lio Ribeiro, Sameer Singh, and Carlos Guestrin.
\newblock Anchors: High-precision model-agnostic explanations.
\newblock In {\em {AAAI}}, pages 1527--1535. {AAAI} Press, 2018.

\bibitem{romera2015embarrassingly}
Bernardino Romera-Paredes and Philip Torr.
\newblock An embarrassingly simple approach to zero-shot learning.
\newblock In {\em International conference on machine learning}, pages
  2152--2161. PMLR, 2015.

\bibitem{selvaraju2017grad}
Ramprasaath~R Selvaraju, Michael Cogswell, Abhishek Das, Ramakrishna Vedantam,
  Devi Parikh, and Dhruv Batra.
\newblock Grad-cam: Visual explanations from deep networks via gradient-based
  localization.
\newblock In {\em Proceedings of the IEEE international conference on computer
  vision}, pages 618--626, 2017.

\bibitem{selvaraju2016grad}
Ramprasaath~R Selvaraju, Abhishek Das, Ramakrishna Vedantam, Michael Cogswell,
  Devi Parikh, and Dhruv Batra.
\newblock Grad-cam: Why did you say that?
\newblock {\em arXiv preprint arXiv:1611.07450}, 2016.

\bibitem{LXMERT}
Hao Tan and Mohit Bansal.
\newblock {LXMERT:} learning cross-modality encoder representations from
  transformers.
\newblock In {\em {EMNLP/IJCNLP} {(1)}}, pages 5099--5110. Association for
  Computational Linguistics, 2019.

\bibitem{van2017neural}
Aaron Van Den~Oord, Oriol Vinyals, et~al.
\newblock Neural discrete representation learning.
\newblock {\em Advances in neural information processing systems}, 30, 2017.

\bibitem{cub}
Catherine Wah, Steve Branson, Peter Welinder, Pietro Perona, and Serge
  Belongie.
\newblock The caltech-ucsd birds-200-2011 dataset.
\newblock 2011.

\bibitem{wang2021average}
Zihan Wang, Chengyu Dong, and Jingbo Shang.
\newblock " average" approximates" first principal component"? an empirical
  analysis on representations from neural language models.
\newblock {\em arXiv preprint arXiv:2104.08673}, 2021.

\bibitem{wei2022masked}
Chen Wei, Haoqi Fan, Saining Xie, Chao-Yuan Wu, Alan Yuille, and Christoph
  Feichtenhofer.
\newblock Masked feature prediction for self-supervised visual pre-training.
\newblock In {\em Proceedings of the IEEE/CVF Conference on Computer Vision and
  Pattern Recognition}, pages 14668--14678, 2022.

\bibitem{xie2022visual}
Yujia Xie, Luowei Zhou, Xiyang Dai, Lu Yuan, Nguyen Bach, Ce Liu, and Michael
  Zeng.
\newblock Visual clues: Bridging vision and language foundations for image
  paragraph captioning.
\newblock {\em arXiv preprint arXiv:2206.01843}, 2022.

\bibitem{yan2022radbert}
An Yan, Julian McAuley, Xing Lu, Jiang Du, Eric~Y Chang, Amilcare Gentili, and
  Chun-Nan Hsu.
\newblock Radbert: Adapting transformer-based language models to radiology.
\newblock {\em Radiology: Artificial Intelligence}, 4(4):e210258, 2022.

\bibitem{L2C}
An Yan, Xin~Eric Wang, Tsu-Jui Fu, and William~Yang Wang.
\newblock L2c: Describing visual differences needs semantic understanding of
  individuals.
\newblock {\em arXiv preprint arXiv:2102.01860}, 2021.

\bibitem{PA3D}
An Yan, Yali Wang, Zhifeng Li, and Yu Qiao.
\newblock Pa3d: Pose-action 3d machine for video recognition.
\newblock In {\em Proceedings of the ieee/cvf conference on computer vision and
  pattern recognition}, pages 7922--7931, 2019.

\bibitem{UniCL}
Jianwei Yang, Chunyuan Li, Pengchuan Zhang, Bin Xiao, Ce Liu, Lu Yuan, and
  Jianfeng Gao.
\newblock Unified contrastive learning in image-text-label space.
\newblock {\em CoRR}, abs/2204.03610, 2022.

\bibitem{yang2022language}
Yue Yang, Artemis Panagopoulou, Shenghao Zhou, Daniel Jin, Chris
  Callison-Burch, and Mark Yatskar.
\newblock Language in a bottle: Language model guided concept bottlenecks for
  interpretable image classification.
\newblock {\em arXiv preprint arXiv:2211.11158}, 2022.

\bibitem{CoCa}
Jiahui Yu, Zirui Wang, Vijay Vasudevan, Legg Yeung, Mojtaba Seyedhosseini, and
  Yonghui Wu.
\newblock Coca: Contrastive captioners are image-text foundation models.
\newblock {\em CoRR}, abs/2205.01917, 2022.

\bibitem{florence}
Lu Yuan, Dongdong Chen, Yi{-}Ling Chen, Noel Codella, Xiyang Dai, Jianfeng Gao,
  Houdong Hu, Xuedong Huang, Boxin Li, Chunyuan Li, Ce Liu, Mengchen Liu,
  Zicheng Liu, Yumao Lu, Yu Shi, Lijuan Wang, Jianfeng Wang, Bin Xiao, Zhen
  Xiao, Jianwei Yang, Michael Zeng, Luowei Zhou, and Pengchuan Zhang.
\newblock Florence: {A} new foundation model for computer vision.
\newblock {\em CoRR}, abs/2111.11432, 2021.

\bibitem{PCBM}
Mert Y{\"{u}}ksekg{\"{o}}n{\"{u}}l, Maggie Wang, and James Zou.
\newblock Post-hoc concept bottleneck models.
\newblock {\em CoRR}, abs/2205.15480, 2022.

\bibitem{compDL}
Tian Yun, Usha Bhalla, Ellie Pavlick, and Chen Sun.
\newblock Do vision-language pretrained models learn primitive concepts?
\newblock {\em arXiv preprint arXiv:2203.17271}, 2022.

\bibitem{zhang2022opt}
Susan Zhang, Stephen Roller, Naman Goyal, Mikel Artetxe, Moya Chen, Shuohui
  Chen, Christopher Dewan, Mona Diab, Xian Li, Xi~Victoria Lin, et~al.
\newblock Opt: Open pre-trained transformer language models.
\newblock {\em arXiv preprint arXiv:2205.01068}, 2022.

\bibitem{zhou2021ibot}
Jinghao Zhou, Chen Wei, Huiyu Wang, Wei Shen, Cihang Xie, Alan Yuille, and Tao
  Kong.
\newblock ibot: Image bert pre-training with online tokenizer.
\newblock {\em arXiv preprint arXiv:2111.07832}, 2021.

\end{thebibliography}
}

\clearpage

\appendix

\section{Prompt Design and Queried Attributes}
\label{app:prompt_design}

\subsection{GPT3}
\noindent Inspired by recent work on querying LLMs~\cite{xie2022visual,pratt2022does},  we start with the following prompt and a demonstration to query GPT3: \\\\
\texttt{Q: What are useful visual features to distinguish a lemur in a photo?\\
A: There are several useful visual features to tell there is a lemur in a photo:\\
- four-limbed primate\\
- black, grey, white, brown, or red-brown\\
- wet and hairless nose with curved nostrils\\
- long tail\\
- large eyes\\
- furry bodies\\
- clawed hands and feet\\
Q: What are useful visual features to distinguish {\textrm{\emph{class\_name}}} in a photo?
A: There are several useful visual features to distinguish {\textrm{\emph{class\_name}}} in a photo:
}\\

\noindent To elicit knowledge within a certain domain, we also test the following prompt to specify the domain given a task:

\texttt{Q: What are useful visual features to distinguish {\textrm{\emph{class\_name}}} from other \textrm{\emph{domain\_name}} in a photo?
A: There are several useful visual features to distinguish {\textrm{\emph{class\_name}}} from other \textrm{\emph{domain\_name}} in a photo:
}\\

Here \textrm{\emph{class\_name}} is the name of each class in the datasets. For instance, in CIFAR-10, \textrm{\emph{class\_name}} is from \texttt{\{airplane, automobile, bird, cat, deer, dog, frog, horse, ship, truck\}}. \emph{domain\_name} is the domain of the datasets. We set \emph{domain\_name} to be \texttt{birds, objects, objects, flowers, foods, dogs and cats, cars, animals} for datasets CUB, CIFAR-10, CIFAR-100, Flowwer, Food, Oxford-pets, Stanford-cars, Imagenet-Animals, respectively.

\subsection{ChatGPT}
\paragraph{Query ChatGPT for CUB}
As CUB is in a specific domain of bird species, we use ChatGPT to design structured and compositional attributes as in~\cite{cub}. 
Specifically, we first query ChatGPT with the following prompts to obtain the possible names describing \emph{body parts} for the birds:\\
\texttt{What are the possible body parts to visually distinguish birds in the photo? }\\
We obtain the following attributes for \emph{body parts}: \\
$\mathcal{BP}=$\texttt{\{wings, beak, feet, tail, head, breast, abdomen, leg, feathers\}}. \\
Then we query for possible colors: \\
\texttt{What are the possible colors that are possible to appear on a bird? }\\
which results in a set of colors: \\
$\mathcal{C}=$\texttt{\{red, orange, yellow, green, blue, purple, brown, black, white, gray\}}. \\
Then we query the shapes of each possible body part, take \texttt{wings} for example:\\
\texttt{What are the possible shapes for bird wings?}
\begin{table}[]
    \centering
    \begin{tabular}{c|p{5.8cm}}
        \toprule
        Body parts & Possible shapes \\
        \midrule
        wings & \texttt{Swallowtail or fork-tailed wings, Round wings, Long, narrow wings, Short, broad wings, Elliptical wings} \\
        \midrule
        beak & \texttt{Conical beaks, Hooked beaks, Probe-like beaks, Wide, flat beaks, Short, stubby beaks, Long, thin beaks}\\
        \midrule
        feet & \texttt{Webbed feet, Talons, Perching feet, Scaling feet, Running feet}\\
        \midrule
        tail & \texttt{Fan-shaped tails, Square-shaped tails, Rounded tails, Forked tails, Tails with streamers}\\
        \midrule
        head & \texttt{Conical heads, Round heads, Elongated heads, Wide heads, Stout heads, Narrow heads} \\
        \midrule
        breast & \texttt{Flat breasts, Round breasts, Bulky breasts, Slender breasts}\\
        \midrule
        leg & \texttt{Long and slender legs, Short and thick legs, Webbed legs, Talons legs, Perching legs}\\
        \midrule
        abdomen & \texttt{Round and plump abdomen, Slim and streamlined abdomen, Long and thin abdomen, Puffed out abdomen}\\
        \midrule
        feathers & \texttt{Long, narrow feathers, Short, broad feathers, Round feathers, Streamer-like feathers}\\
        \bottomrule
    \end{tabular}
    \caption{Possible shapes for each body part of birds.}
    \label{tab:queried_possible_shapes_for_each_body_part_of_birds}
    \vspace{-10pt}
\end{table}

Finally, with the colors $\mathcal{C}$ and the possible shapes for each body part shown in Table \ref{tab:queried_possible_shapes_for_each_body_part_of_birds}, we build 440 attributes for CUB with examples shown below: \\
\texttt{
red swallowtail or fork-tailed wings. \\
red round wings. \\
green webbed legs. \\
orange round wings. \\
}
\vspace{-10pt}
\paragraph{Query ChatGPT for CIFAR-100}
We utilize the batch prompting method described in Section \ref{sec:methodology} to query ChatGPT for the attributes of CIFAR-100. 

\noindent We use the following prompt to query ChatGPT: \\\\
\texttt{Q: Here are five {\textrm{\emph{superclass\_name}}}: {\textrm{\emph{\{class\_name\_1,...,class\_name\_N\}}}}. What are the useful visual features for distinguishing them in a photo? Please list every attribute in bullet points.\\
}\\
Here \textrm{\emph{superclass\_name}} is the name of each superclass in the datasets. For instance, in CIFAR-100, beaver, dolphin, otter, seal, whale belongs to the superclass aquatic mammals.

\subsection{Comparing Different Attribute Concept Pools}
% Yu
With the above prompting templates, we explore the effects of different concept pools. Comparing with the concept pool constructed from GPT-3 prompts with corresponding class names for each dataset, we add the following two pools to discuss the effects: (1) the concepts from Imagenet queried from GPT-3, which would be larger and also noisier; (2) The concepts from ChatGPT. 
Note that we conduct the ablation study on two dataset CUB and CIFAR-100, since their classes are generally covered by the classes from Imagenet. For CUB, we manually designed the attributes in the pool and for CIFAR-100, the attributes are queried in a hierarchical way (see Appendix \ref{app:prompt_design} for the details). The results are shown in Table \ref{tab:variation_of_concept_pool}. 

Overall, our learning-to-search method is robust to different attribute pools, and we do not observe significant performance change using GPT-3 or ChatGPT. Though our human-designed compositional attributes with ChatGPT on CUB is worse than pure LLM-generated attributes.

\begin{table}[]
    \centering
    \resizebox{\linewidth}{!}{%
    \begin{tabular}{c|ccc|cccc}
        \toprule
        Datasets & \multicolumn{3}{|c|}{CUB} & \multicolumn{3}{c}{CIFAR-100} \\ 
        \midrule
        K & 8 & 16 & 32 & 8 & 16 & 32 \\
        \midrule
        GPT-3 & \textbf{31.67} & 48.55 & 60.27 & \textbf{34.77} & \textbf{52.24} & 66.30 \\
        GPT-3-Imagenet & 30.81 & \textbf{49.29} & \textbf{60.41} & 33.80 & 51.01 & 65.61 \\
        ChatGPT & 21.66 & 40.28 & 47.46 & 
33.79 & 51.26 & \textbf{67.06} \\
        \bottomrule
    \end{tabular}}
    \caption{Comparison \emph{w.r.t.} different Concept Pools.}
    \label{tab:variation_of_concept_pool}
\end{table}

\subsection{Robustness Check}
To confirm the effectiveness of our prompts and the robustness of GPT-3 prompts, we conduct the experiments with the concepts queried from GPT3 using different prompts. We design semantically instructive and misleading prompts as shown in Table \ref{performances_with_different_prompts}. Overall, we observe that other instructive prompts perform similar to ours, while misleading prompts could hurt the performance drastically.

\begin{table*}[]
    \centering
    \begin{tabular}{c|p{14cm}}
        \toprule
 Category &  \begin{tabular}[c]{p{0.9\linewidth}|c} \begin{tabular}[c]{@{}c@{}} Prompts \end{tabular} & Acc \end{tabular} \\
 \midrule
 \begin{tabular}[c]{@{}c@{}} Instructive \end{tabular} & 
\begin{tabular}{p{0.9\linewidth}|c}
\begin{tabular}{p{\linewidth}}
\texttt{What are the useful visual features to distinguish {\textrm{\emph{class\_name}}}?}
\end{tabular} & 31.67\\
\midrule
% \midrule
\begin{tabular}{p{\linewidth}}
\texttt{What are the helpful visual features to distinguish {\textrm{\emph{class\_name}}}?}
\end{tabular}
& \begin{tabular}[c]{@{}c@{}} \textbf{32.71} \end{tabular}\\
\midrule
\begin{tabular}{p{\linewidth}}
\texttt{What are the distinctive visual features to distinguish {\textrm{\emph{class\_name}}}?}
\end{tabular}
& \begin{tabular}[c]{@{}c@{}} 30.38 \end{tabular}\\
\end{tabular}   \\
        \midrule
\begin{tabular}[c]{@{}c@{}} Misleading \end{tabular} & 
\begin{tabular}{p{0.9\linewidth}|c}
\begin{tabular}{p{\linewidth}}
\texttt{What are the useless visual features to distinguish {\textrm{\emph{class\_name}}}?}
\end{tabular} & 19.64\\
\midrule
% \midrule
\begin{tabular}{p{\linewidth}}
\texttt{Give me some random visual features in a photo to distinguish {\textrm{\emph{domain\_name}}}:}
\end{tabular}
& \begin{tabular}[c]{@{}c@{}} 5.85 \end{tabular} \\
\end{tabular}
\\
\bottomrule
\end{tabular}
\caption{Robustness study against different prompts on CUB, with $K=8$}
\label{performances_with_different_prompts}
\end{table*}

\section{Implementation Details}
\label{app:implementation_details}
% \subsection{Regularization Design and Hyperparameter Tuning}
% Then we calculate the mean distance of each concept concept embedding $T_i, i\in\{1,\cdots,N\}$ to the distribution defined by $\boldsymbol{\mu}$ and $\textbf{S}$ as:
% \begin{equation}
%     \overline{\mathcal{D}}_{\mathit{mah}} = \sum_{i=1}^N \sqrt{(\textbf{T}_i - \boldsymbol{\mu}) \textbf{S}^{-1} (\textbf{T}_i - \boldsymbol{\mu})}
% \end{equation}

% With this average distance, we define the following loss function:
% \begin{equation}\label{eq:mahalanobis_expanded}
%     \mathcal{L}_{\mathit{mah}}^j (\lambda) = \frac{\mathcal{D}_{\mathit{mah}}^j - \overline{\mathcal{D}}_{\mathit{mah}}}{(\overline{\mathcal{D}}_{\mathit{mah}})^\lambda}
% \end{equation}
% where $\lambda$ is the hyperparamter to be tuned. 
% \subsection{Dataset Statistics}

\subsection{Linear Probing}
After we obtain attribute embeddings $\textbf{T}^*$ from Eq.(\ref{eq:text_embedding_searching}), we then calculate the semantic vector $\textbf{A}^*$ of each image $I$ with the D-dimensional image embedding $\textbf{V} = \Theta_V(I) \in \mathbb{R}^D$: 
\begin{align}
    s_j &= \cos (\textbf{V}, \textbf{T}^*_j), j=1,\cdots,K, \\
    \mathbf{A}^* &= \Trans{(s_1,\cdots,s_K)}.
\end{align}
Then we calculate the score vectors of all the images in the training and testing dataset. We then use linear probing to evaluate the performance. Since we use a task-guided searching during the first stage to find $K$ attribute embeddings, we can readily use the classification head in the first stage (i.e., a linear model $f_\theta \in \mathbb{R}^K \rightarrow \mathbb{R}^{K_C} $ with one fully connected layer) for our second stage with lightweight fine-tuning instead of training from scratch. where $K_C$ is the number of classes. Then, we train $f_\theta$ with a cross-entropy loss:
\begin{equation}\label{eq:loss_func_cross_entropy}
    \mathcal{L} = -\frac{1}{M} \sum_{i=1}^M \sum_{c=1}^{K_C} y_{i,c} \log p_{i,c}',
\end{equation}
the same as Eq.(\ref{eq:loss_cc}), $M$ is the number of images in a mini-batch. $y_{i,c}$ is the binary indicator of $i$-th image the mini-batch belonging to class $c$, and $p_{i,c}$ is the predicted probability of the $i$-th image belonging to class $c$. Then $p_{i,c}', c\in\{1,\cdots,K_C\}$ is calculated as:
\begin{equation}
    \Trans{[p_{i,1}', \cdots, p_{i,K_C}']} = \textrm{Softmax}(f_\theta(\mathbf{A}^*_i))
\end{equation}
where $\textbf{A}^*_i$ is the semantic vector of the $i$-th image in the mini-batch. Then we will use $f_\theta$ to classify the images in the test set to yield the performances.

\section{Additional Experiments}

\subsection{Comparison with Zero-Shot Classifications}
\begin{table}[h!]
\vspace{-10pt}
    \centering
    \resizebox{0.8\linewidth}{!}{%
    \begin{tabular}{c cccc}   
        \toprule
        Datasets & CIFAR-100 & Stanford-cars \\ 
        \midrule
        CLIP-ZS w/ class names & 54.49 & 57.87\\
        CLIP-ZS w/ attributes  & 30.07 & 5.42 \\
         \midrule
         CLIP-Train Visual  & 79.30 & 79.95\\
         Ours (K=512) & 75.41 & 74.67\\
         \midrule
         Datasets & Flower & Imagenet-Animals \\ 
        \midrule
        CLIP-ZS w/ class names & 60.19 & 59.44\\
        CLIP-ZS w/ attributes & 9.80 & 9.13\\
         \midrule
         CLIP-Train Visual & 92.35 & 75.31 \\
         Ours (K=512) & 90.29 & 75.60 \\
         \bottomrule
    \end{tabular}}
    \vspace{-10pt}
    \caption{Comparison with zero-shot classification methods.}
   \vspace{-15pt} \label{tab:comparison_with_zero_shot_clip}
\end{table}

We deliver more results in Table~\ref{tab:comparison_with_zero_shot_clip}. We use \textit{A photo of} as the prompt for all methods. Zero-shot (CLIP-ZS) is worse than supervised training. Note that CLIP-ZS with class names may not be a fair comparison, \textbf{as our goal is to classify images with attributes instead of class names}, thereby gaining a level of interpretability and fine-grained understanding of visual recognition. If we use only attributes for CLIP-ZS, the performance drastically decreases.

\subsection{Human Evaluation}
To further evaluate the quality of our learned attributes, we conduct a pairwise human evaluation on Amazon Mechanical Turk. Specifically, we compare our attributes with uniformly sampled attributes from the GPT-3 generated attributes, and ask human to decide which set of attributes are better. Since datasets with hundreds of classes are hard to reason and compare, we evaluate our results on CIFAR-10.  We sample 100 sets of 4 attributes from the attribute pool, and create 100 pairs of each random set of 4 attributes with our learned 4 attributes. Each pair was assigned to 5 workers to eliminate human variance. 
For each attribute pair, workers are presented with sampled images from CIFAR-10. We instruct the workers to consider which set of attributes are more useful to classfy the 10 classes. As shown in~\cref{tab:human_evaluation_cifar10}, even though in most cases, the attributes look similar to human, workers still favor \textit{Our method} over \textit{Uniform sampling}, which is consistent with the classification accuracy.

\begin{table}
\small
\centering
\begin{tabular}{c c|c|c}
% \toprule
%   & \multicolumn{3}{c}{L2C vs CNN+LSTM}\\
\toprule
Choice (\%) & Ours & Uniform & Tie \\
\midrule
Score & 31.6 & 19.0 & 49.4 \\
\bottomrule
\end{tabular}
\caption{Human evaluation results on CIFAR-10. Human are asked to vote which attributes are better, where \textit{tie} means the two sets looks the same to annotators}
\label{tab:human_evaluation_cifar10}
\end{table}

% \paragraph{Implementation Details}
% For the concept pools queried from GPT3, the sizes are different across datasets. Specifically, The concept pools of 
% In our method, there are two stages of training. The first stage consists of two FC layers, with a layer of batchnorm added in the middle. We only use batchnorm when the training is hard to converge without it (\ie, it may take more than 5000 epochs to converge). After doing some check, we decide to use batchnorm between two FC layers in all datasets except for CUB, CIFAR-10 and Flower. Then for the baselines, the model structure is set to be (batchnorm +) FC where the existence of batchnorm is corresponding to the above descriptions. For the mahalanobis distance, the parameter $\lambda$ in Eq.(\ref{eq:mahalanobis}) is tuned in \{0, 1, 2, 3\}. We also add another parameter setting as setting $\mathcal{L}_{loss} = \mathcal{L}_{ce}$ in Eq.(\ref{eq:loss_func}), \ie, removing the regularization term. The batchsize is set to 4096 for all datasets except that the batchsize for Imagenet\_Animals is set to 32768. We set number of epochs to 5000 epochs while the training is stopped if the best accuracy does not improve over consecutive 100 epochs. The learning rate is set to 0.01 in all experiments.  

\section{More ablations}
\label{app:ablation}
\paragraph{Better V\&L models}
% An
We evaluate different variants of CLIP style models, as shown in~\cref{tab:different_VLM}. Overall, our method is model-agnostic. It can be applied with any VLMs that compute image-text similarities. We also observe that in general, a stronger VLM will result in more accurate estimation of semantic vectors, hence improves classfication performance.

\begin{table}[t]
    \centering
    \resizebox{0.95\linewidth}{!}
    {%
    \begin{tabular}{l|ccc}
    \toprule
        % & \multicolumn{3}{c}{K}\\
        % \midrule
        Model Architectures & 8 & 16 & 32 \\
        \midrule
        CLIP RN50 & 24.28 & 38.95 & 56.11 \\
        CLIP RN101 & 27.36 & 46.10 & 56.06 \\
        CLIP RN50x16 & 28.91 & 53.46 & 57.69\\
        CLIP ViT-B/32 & 31.67 & 48.55 & 60.27 \\
        CLIP ViT-B/16 & 36.69 & 55.64 & 63.70 \\
        CLIP ViT-L/14  & 38.71 & 65.52 & 74.99 \\
        CLIP ViT-L/14@336px  & 40.95 & 66.58  & 76.04 \\
        Open-CLIP ViT-H-14 LAION-2B & 49.84 & 73.28 & 82.60\\
        \bottomrule 
    \end{tabular}}
    \caption{Ablation study on different VLMs with bottleneck size $K$=8,16,32 on the CUB dataset.}
\label{tab:different_VLM}
\end{table}

\begin{figure*}[t]
\subfigure[Long Tailed Jaeger]
{\label{fig:long_tailed_jaeger}\includegraphics[width=0.475\linewidth]{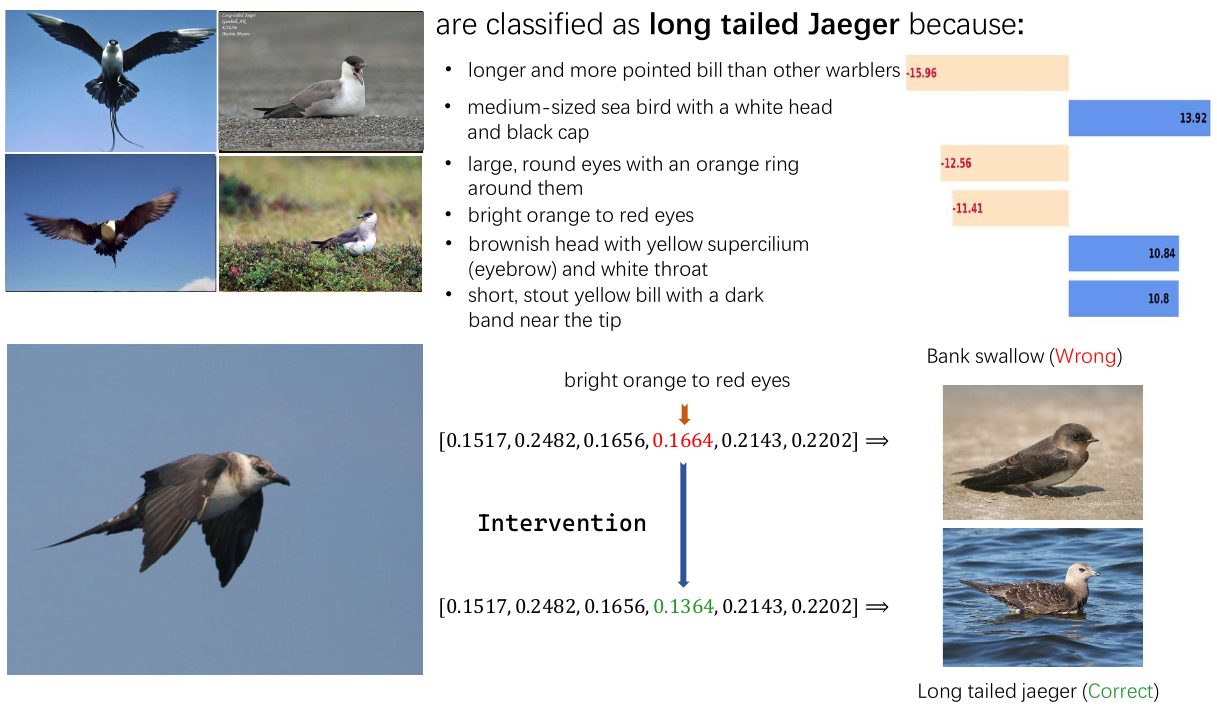}}
\subfigure[White Throated Sparrow]{\label{fig:white_throated_sparrow}\includegraphics[width=0.475\linewidth]{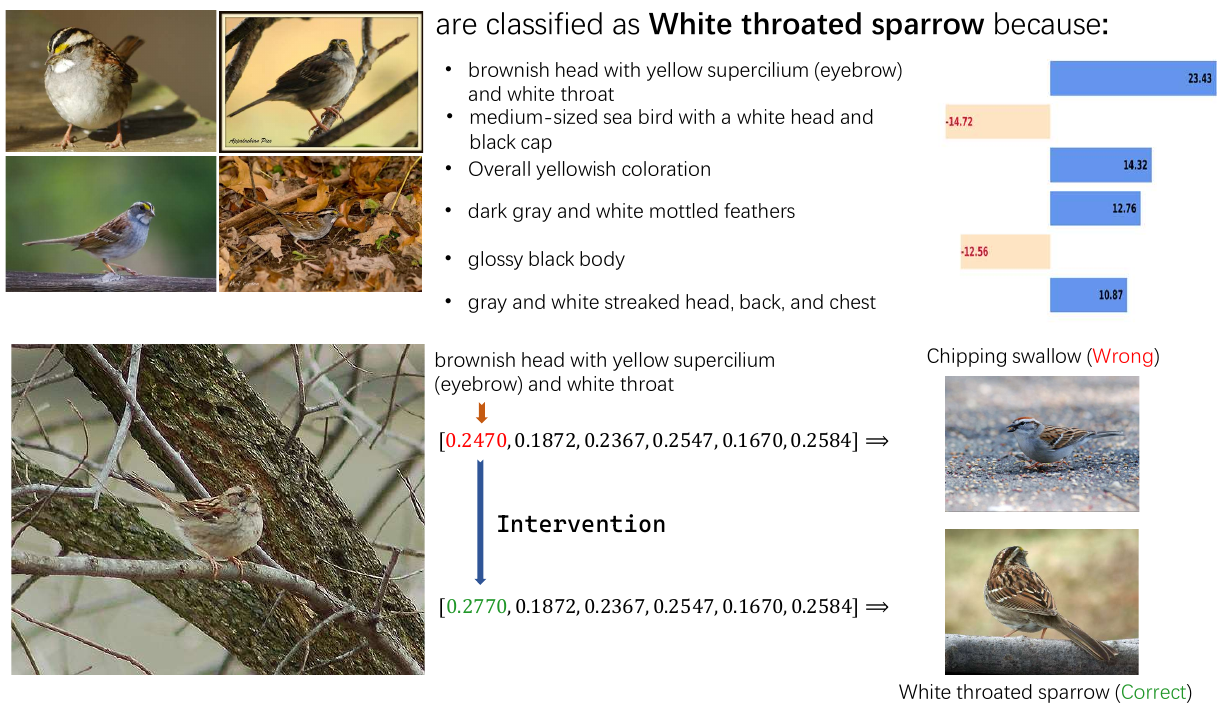}}
\caption{Examples on interpretability and interactivity. (1)~The upper half of each figure show important attributes for two classes of birds. We choose 6 out of 32 attributes with highest importance scores, which are computed by multiplication between clip scores and weights in the linear probe, defined in~\cref{eqn:importance_score}. 
(2)~The lower half of each figure demonstrates the intervention on the semantic vector (i.e., CLIP scores) to correct the prediction, we use $\delta$=0.03 for all interventions on clip scores as an empirical value. The array of 6 scores are of the same order as the attributes.}
\label{fig:test_time_intervene_more}
\end{figure*}

\vspace{-5pt}
\paragraph{Effectiveness of semantic projection. Scores vs one-hot}
% Yu
In this part, we consider the following baseline to show that the information within the similarity scores are useful: For an image $I$, after calculating all the similarity scores between every attribute $a_i \in \{a_1, \cdots, a_N\}$ and the image $I$ to obtain the vector $\mathbf{A} \in \mathbb{R}^N$. Then we wipe out the information in the scores by setting top-$K$ large scores in $\mathbf{A}$ as 1, and setting the left scores as 0, which will give us a binary vector $\mathbf{A}_{bin} \in \{0, 1\}^N $. Then we train and test the classification model on the corresponding binary vectors to compare with our methods. We conduct the ablation study on CUB as an example. We choose $K=8$ and $K=16$ for the comparisons. 

From the results in Table \ref{tab:comparision_between_scores_and_onehot}, we observe the information from the similarity scores provides information for classification, while removing the information in the scores (converting top-K scores to 1) may lead to performance drop. 

\begin{table}[t]
    \centering
    \resizebox{\linewidth}{!}{%
    \begin{tabular}{c|ccc|ccc}
    \toprule
        Datasets & \multicolumn{3}{|c|}{CUB} & \multicolumn{3}{c}{CIFAR-100} \\
        \midrule
        \# of Non-zeros & 8 & 16 & 32 & 8 & 16 & 32 \\
        \midrule
        One-Hot & 36.22 & 44.35 & 48.96 & 57.90 & 61.85 & 65.43 \\
        Only Top Scores & 36.03  & 44.32 & 49.43 & 58.32 & 61.90 & 65.86 \\
        \bottomrule 
    \end{tabular}}
    \caption{Comparison between scores and one-hot}
    \vspace{-15pt}
\label{tab:comparision_between_scores_and_onehot}
\end{table}

\paragraph{GPT-3 attributes vs. Random words}
We present more results on 8 datasets to verify that a large number of GPT-3 attributes behaves similar as random words. The observations are coherent on all eight datasets. When $K$ is large, even if we randomly create $K$ meaningless phrases from the entire vocabulary, we can still obtain competitive classification performance. 

\begin{table}[t]
    \centering
    \resizebox{\linewidth}{!}{%
    \begin{tabular}{ccccc}
    \toprule
         & \multicolumn{2}{c}{Nonsense} & \multicolumn{2}{c}{GPT3} \\
         \midrule
         $K$ & 4 & 512 & 4 & 512 \\
         \midrule
        CUB & 2.42 & 64.79 & \textbf{12.98} & 67.64  \\
        CIFAR-10 & 31.26 & 92.81 & \textbf{60.30} & 93.67 \\
        CIFAR-100 & 10.17 & 77.40 & \textbf{16.13} & 77.55 \\
        Flower & 3.33 & 90.20$^*$ & \textbf{28.92} & 90.78$^*$ \\
        Food &  7.79 & 82.65$^*$ & \textbf{16.23} & 82.50$^*$ \\
        Oxford\_pets & 14.31 & 83.61$^*$ & \textbf{28.07} & 86.01$^*$ \\
        Stanford\_cars & 5.06 & 75.09 & \textbf{13.41} & 75.13 \\
        Imagenet\_Animals & 3.78 & 75.12 & \textbf{8.81} & 75.75 \\
        \bottomrule
    \end{tabular}
    }
    \caption{GPT3 vs Nonsense. $*$ means the results are obtained when setting the number of attributes to the size of the concept pool from GPT3, which corresponding to the results of "full" in Figure \ref{fig:overall_performances}. }
    \label{tab:gpt3_vs_nonsense}
\end{table}

\section{Additional Case Study}
\label{app:case_study}

\subsection{Test-time Intervention}
We provide more case studies in~\cref{fig:test_time_intervene_more} to show the interpretability and interactivity of our method. 
% Similar to~\cref{fig:test_time_intervene}, averaged importance scores of a class are shown in the top, with intervention result on a single image shown in the bottom.

\subsection{Visualization of Discovered Attributes}
\begin{table*}[]
    \centering
    \begin{tabular}{c|p{14cm}}
        \toprule
Dataset & Learned 32 attributes for each dataset \\
        \midrule
\begin{tabular}[c]{@{}c@{}} CUB \end{tabular} & 
\begin{tabular}[c]{p{\linewidth}}{{\footnotesize(1) Brown, gray and white feathers on the upper parts of the body, with a rusty red or pinkish tinge to the head; (2) bright yellow and black coloring; (3) distinctive white throat; (4) Broad tail that is shorter than other pelican species ; (5) short legs for perching in trees ; (6) bright yellow throat, breast, and flanks with black bars ; (7) Brown and white mottled back ; (8) pinkish red breast patch with white edges ; (9) white throat patch bordered by black stripes ; (10) unique pattern of spots on lower throat and breast.; (11) Large feet for scratching in leaf litter; (12) brownish head with yellow supercilium (eyebrow) and white throat ; (13) red or orange coloration; (14) iridescent black body with blue and purple highlights ; (15) red, black and white feathers; (16) grayish brown body with darker wings and tail; (17) Heavy bill for crushing seeds ; (18) olive green back and wings; (19) white throat, belly and wing bars  ; (20) Long, slender bill with yellow tip; (21) large, white bird; (22) grayish brown head and back; (23) red/orange coloration on the face during breeding season; (24) Gray head and yellow throat with white eye rings; (25) barred wings and tail feathers in black, white and grey patterning ; (26) distinctive white throat patch ; (27) long, heavy bill; (28) yellow head; (29) male mallards have a green head, yellow beak and white neck ring; (30) Broad white eyering ; (31) white throat and belly region; (32) bright orange and black plumage}}\end{tabular}\\
\midrule
\begin{tabular}[c]{@{}c@{}} CIFAR-10 \end{tabular} & 
\begin{tabular}[c]{p{\linewidth}}{\footnotesize(1) antlers (in males); (2) some birds have crests on their heads; (3) propellers or jet engines; (4) fur coat of varying colors and patterns; (5) a tail with a horizontal stabilizer; (6) portholes along the hull; (7) large body with a cab and a bed; (8) landing gear; (9) four wheels; (10) fuselage and other structural elements; (11) four-wheeled vehicle; (12) long head with a mane and tail; (13) has a mast with sails or flags; (14) paws with clawed toes; (15) feathers of various colors and patterns; (16) tail lights; (17) furry body; (18) a beak or bill for eating, preening, and other activities; (19) pointed bow and stern; (20) mane of hair along the neck and back; (21) masts, sails, and rigging; (22) two wings and two legs; (23) moist slimy skin; (24) windshield and side windows; (25) rudders at the stern for steering; (26) smokestacks or funnels on top of the ship; (27) a large deck or superstructure; (28) hooves on each foot; (29) typically has a steering wheel and pedals for driving; (30) control surfaces (flaps, ailerons, rudder); (31) grille or front fascia; (32) may have a cargo area in the back}\end{tabular}\\
\midrule
\begin{tabular}[c]{@{}c@{}} CIFAR-100 \end{tabular} & 
\begin{tabular}[c]{p{\linewidth}}{\footnotesize(1) A pair of pedipalps near the mouth used for sensing, holding prey and mating; (2) multiple petals in shades of pink, red, yellow or white; (3) Headboard and footboard; (4) pouch on the abdomen of female kangaroos; (5) fruits are two winged samaras in clusters; (6) large courtyard area surrounded by buildings; (7) orange or yellow fur with black stripes; (8) buds clustered at branch tips in winter months; (9) large wheels and tires; (10) Mattress and bedding; (11) large, floppy ears; (12) silver scales with black and red spots on the sides; (13) long snout with sharp teeth; (14) smooth oval shaped sepals; (15) clustered, coconuts at the tip of each branch; (16) white foam from waves breaking against rocks and shorelines; (17) designs, colors, or patterns on the can; (18) shaggy fur; (19) tailgate at rear end; (20) long bushy tail usually with a tuft of hair at the end; (21) portcullis at entrance to gatehouse; (22) cabin or operator’s seat in the middle of the vehicle; (23) drawbridge over a moat; (24) dialing pad with numbers 0–9; (25) catkins (flowers) in spring; (26) a large, tawnycolored body with a shaggy mane; (27) Ten walking legs and two large antennae; (28) Rim around the edge of the bowl; (29) armrests and backrests; (30) Long stem or pole extending from the shade to the base; (31) an ovary located at the base of the flower; (32) waxy texture of the petals and leaves}\end{tabular}\\
\midrule
\begin{tabular}[c]{@{}c@{}} Flower \end{tabular} & \begin{tabular}[c]{p{\linewidth}}{\footnotesize(1) bright purple petals that are fused together to form a thistle-like shape; (2) trumpet shaped flowers in shades of blue, purple, and white; (3) blue, purple, or white flowers with a thistle-like appearance; (4) an upright inflorescence (flower spike) bearing several clustered flowers on each branch ; (5) umbel of several small flowers on top of a single stem ; (6) Center of the spathe that looks like a tail or spadix; (7) the flower is a daisy-like plant with white petals and yellow center; (8) layered petals with a yellow center and pink edges ; (9) Large, bright pink to red flower ; (10) Six distinct petal segments surrounding an inner cup of short filaments and a trumpet center.; (11) trumpet shaped orange center with yellow stamens protruding from it ; (12) large, white petals with a yellow center; (13) fragrant single or double blooms in white, pink, or red; (14) brightly colored petals in shades of oranges, reds, and yellows; (15) large, yellow petals that form a daisy-like shape; (16) umbrella shaped clusters of white to pinkish flowers ; (17) tall, leafless stem; (18) hibiscus shaped leaves that are serrated around the edges; (19) pink to purple colored petals with red lips; (20) single stem with a rosette of leaves; (21) tall, slender stem with a single umbel of flowers; (22) large blue or purple flowers with five petals and a hooded center; (23) intricate patterns of blue, purple, pink and white lines on the petals ; (24) yellowish green sepals below the flower; (25) dark purple petals; (26) five petals arranged around a central column of white stamens and stigma ; (27) pink, white, or lavender flowers with five petals; (28) bright yellow flower head; (29) bright pink, red, or white petals with fringed edges; (30) bright red, orange, or yellow blooms; (31) bright red and yellow petals; (32) deep red, orange or yellow petals}\end{tabular}\\
\bottomrule
    \end{tabular}
    \caption{Learned 32 attributes on CUB, CIFAR-10, CIFAR-100 and Flower.}
    \label{tab:pruned_features_with_K32}
\end{table*}

\begin{table*}[]
    \centering
    \begin{tabular}{c|p{14cm}}
        \toprule
        Dataset & Learned 32 attributes for each dataset\\
        \midrule
\begin{tabular}[c]{@{}c@{}} FOOD \end{tabular} & 
\begin{tabular}[c]{p{\linewidth}}{\footnotesize(1) sliced strawberries arranged over the cream/whipped topping; (2) A white or wheat bun with a golden brown exterior ; (3) Long, thin rice noodles; (4) lattice pattern on the top layer made by weaving strips of pastry dough; (5) slices of apples arranged in a spiral pattern ; (6) large pieces of clams visible in the chowder; (7) Gyoza is typically shaped like a half-moon or dumpling and can have either open or closed tops. ; (8) The broth will usually have either a sour or spicy taste depending on the type of soup ; (9) red sauce layered between the noodles and cheese; (10) shredded carrots embedded within the cake ; (11) dollop of sour cream or guacamole ; (12) thin layers of phyllo dough; (13) tender squid rings inside ; (14) distinctive pattern of takoyaki sauce on top; (15) moist and dark brown cake with visible cocoa powder; (16) Served on top of a bed of shredded daikon radish or grated daikon ; (17) lobster chunks mixed with mayonnaise and spices; (18) two layers of toast with lettuce, bacon, and tomatoes in between; (19) mashed avocado texture ; (20) cooked shrimp in a variety of colors (pink, orange, etc.); (21) steaming bowl of soup with steam rising up ; (22) melted cheese over the chips; (23) The presence of Mandarin pancakes, cucumber slices and spring onion used in traditional preparation methods; (24) fried or served cold with dipping sauce; (25) a custard base topped with a layer of hardened caramelized sugar; (26) a dollop of gochujang (red pepper) paste ; (27) butter or oil is used to toast the bread on both sides ; (28) chunks of vegetables, tofu, and seaweed floating in it; (29) gooey mixture of sugar, butter and cinnamon visible between the layers of apple slices; (30) olive oil and soy sauce dressing ; (31) toppings such as egg, vegetables, seaweed, and pork slices ; (32) toppings such as jalapenos, tomatoes, onions and/or peppers}\end{tabular}\\
\midrule
\begin{tabular}[c]{@{}c@{}} Oxford-pets \end{tabular} & 
\begin{tabular}[c]{p{\linewidth}}{\footnotesize(1) Long legs and neck; (2) large upright ears; (3) “Ragdoll” appearance with a long body and short legs; (4) Soft wiry coat in black or brindle colors; (5) Pointed ears; (6) Shade of red or wheaten color; (7) dark brown or black coat with white markings; (8) Markings resembling a leopard or tiger in various colors (brown, black, white, orange); (9) mediumsized dog; (10) greyish blue fur with silver tips; (11) short, almost hairless body with wrinkles; (12) Round eyes in shades of blue or green; (13) White and grey fur; (14) A tail that curls over its back; (15) Ears that are small and rounded at the tips; (16) loose skin on the face and neck that can create wrinkles; (17) triangular ears; (18) white blaze on face and chest; (19) droopy ears that hang close to the head; (20) wide eyes with prominent wrinkles around them; (21) thick, white double coat; (22) foxy head and face with a curled tail; (23) Curly tail that curls over the back; (24) black face mask on white fur background; (25) long, silky coat in white or white and black colors; (26) thick mane around neck and chest; (27) distinctive wrinkles on the face; (28) short coat of glossy black fur; (29) Short, glossy coat of black and silver; (30) double coat of fur that is typically fawn, black or silver; (31) black and tan coloring; (32) Visible spots on the body}\end{tabular}\\
\midrule
\begin{tabular}[c]{@{}c@{}} Stanford-cars \end{tabular} & 
\begin{tabular}[c]{p{\linewidth}}{\footnotesize(1) large grille with a classic Bentley badge; (2) Signature wheel arches; (3) large tailgate spoiler on the liftgate; (4) unique wheels with five spokes and silver finish; (5) Front bumper has a skid plate design; (6) flared wheel arches that give the car an aggressive look; (7) Ron Fellows Edition badge on the rear of the car; (8) The distinct hexagonal grille with the Volvo emblem at the center; (9) Distinctive red and black racing stripes with Abarth logo; (10) “4Runner” badge on the rear liftgate; (11) Hatchback style trunk/boot area; (12) High performance tires with "Type R" on the sidewall; (13) horizontal three bar tail lamps with the running Mustang logo at its center.; (14) distinctive grille with a mesh pattern and Spyker logo; (15) Interior: Leather wrapped steering wheel with audio controls; (16) kidney grille with large blue and white BMW logo; (17) black power convertible top; (18) Red badge with "Integra Type R" logo or lettering on the hood and trunk lid; (19) Chrome grille with the Chevrolet logo; (20) the Fiat logo on the front grille and rear of car; (21) Quattro badge on the rear right side of the car; (22) gloss black paint job with distinct yellow detailing; (23) LED tail lights with a unique curved design to give it a modern look.; (24) Chrome grille with the Chrysler emblem in the center; (25) chrome grille with the Chevrolet logo; (26) wide grille with a large chrome Bentley badge in the center; (27) two door hardtop convertible body style; (28) distinctive side windows with curved lines and signature Maybach logo; (29) tailgate spoiler on the rear hatchback door; (30) Chrome grille outlining the Honda logo; (31) Black brake calipers with Corvette lettering; (32) Twodoor, fourseater convertible hardtop}\end{tabular}\\
\midrule
\begin{tabular}[c]{@{}c@{}} Imagenet-Animals \end{tabular} & 
\begin{tabular}[c]{p{\linewidth}}{\footnotesize(1) smaller and lighter than other Welsh corgis; (2) from Airedale, England; (3) male rams have large, thick horns, while female rams have smaller, thinner horns; (4) dense, flat coat; (5) the Maltese has a reputation for being lively, playful and affectionate; (6) coat is predominantly black and tan; (7) Gordon setters are typically black with tan markings; (8) Saint Bernards are large dogs; (9) Kerry blue terriers are from Ireland; (10) the breed name (elkhound); (11) English setters are bred in England; (12) shaggy, matted coat; (13) all black coat; (14) spotted or striped fur; (15) dark brown or black coat with a distinctive "water spaniel" curl; (16) glossy black feathers with a green or blue sheen; (17) pink, orange, or yellow stripes on the shell; (18) black and white, blue and white, or wheaten (red) coloration; (19) wrinkles on the face and head; (20) coat is wheaten in color (ranging from pale cream to rich gold); (21) big ears; (22) male finches have a bright red breast; (23) creamy white or wheaten-colored coat; (24) white throat; (25) large, broad carapace; (26) dark plumage with black and iridescent blue feathers; (27) long, black antennae; (28) large, black and white dolphin; (29) longer legs than other hound breeds; (30) fawn or brindle coloration; (31) long, wirehaired coat; (32) fawn to mahogany coat}\end{tabular}\\
\bottomrule
    \end{tabular}
    \caption{Learned 32 attributes on Food, Oxford-pets, Stanford-cars and Imagenet-Animals.}
    \label{tab:pruned_features_with_K32_on_last_four_datasets}
\end{table*}

We present our learned 32 attributes for each dataset (by setting $K=32$) in~\cref{tab:pruned_features_with_K32} and~\ref{tab:pruned_features_with_K32_on_last_four_datasets}.  Similar to~\cref{fig:pruned_features}, we can observe these attributes are distinctive within each domain, and provides fine-grained attributes to summarize a dataset. 
To some level, we can view these automatically learned attributes as a form of knowledge to help humans understand how visual recognition works.

\end{document}